\documentclass[journal]{IEEEtran}

\usepackage{times}
\usepackage{epsfig}
\usepackage{graphicx}
\usepackage{amsmath}
\usepackage{amssymb}
\usepackage{multirow}
\usepackage{cite}
\usepackage{array}
\usepackage{xcolor}
\usepackage{float}
\usepackage{url}
\usepackage{soul}
\usepackage{amsfonts}
\usepackage{mathrsfs}

\ifCLASSINFOpdf
\else
\fi

\begin{document}

\indent

© 2022 IEEE. Personal use of this material is permitted. Permission from IEEE must be obtained for all other uses, in any current or future media, including reprinting/republishing this material for advertising or promotional purposes, creating new collective works, for resale or redistribution to servers or lists, or reuse of any copyrighted component of this work in other works. Digital Object Identifier 10.1109/TMM.2022.3154600

\title{VCGAN: Video Colorization with Hybrid Generative Adversarial Network}

\author{Yuzhi~Zhao,~\IEEEmembership{Graduate~Student~Member,~IEEE,}
        Lai-Man~Po,~\IEEEmembership{Senior~Member,~IEEE,}
        Wing-Yin~Yu,~\IEEEmembership{Graduate~Student~Member,~IEEE,} Yasar Abbas Ur Rehman,~\IEEEmembership{Member,~IEEE,} Mengyang~Liu, Yujia~Zhang, Weifeng~Ou% <-this % stops a space
\thanks{Manuscript received October 15, 2020; revised April 29, 2021; revised December 1, 2021; accepted February 20, 2022. \textit{(Corresponding author: Yuzhi Zhao.)}}
\thanks{Y. Zhao, L.-M. Po, W.-Y. Yu, and Y. Zhang are with the Department of Electronic Engineering, City University of Hong Kong, Hong Kong, China (e-mail: yzzhao2-c@my.cityu.edu.hk; eelmpo@cityu.edu.hk; wingyinyu8-c@my.cityu.edu.hk; yzhang2383-c@my.cityu.edu.hk).}% <-this % stops a space
\thanks{Y.-A.-U. Rehman is with TCL Corporate Research Hong Kong, Hong Kong, China (e-mail: yasar.abbas@my.cityu.edu.hk).}
\thanks{M. Liu is with Tencent Video, Tencent Holdings Ltd, China (e-mail: mengyaliu7-c@my.cityu.edu.hk).}
\thanks{W. Ou is with SenseTime Group Limited, Hong Kong, China (e-mail: weifengou2-c@my.cityu.edu.hk).}
}% <-this % stops a space}

% The paper headers
\markboth{IEEE Transactions on Multimedia}%
{Shell \MakeLowercase{\textit{Zhao et al.}}: VCGAN: Video Colorization with Hybrid Generative Adversarial Network}

%%%--------------------------------
%%% Proposed by Yuzhi:
% 1. Please delete a content by using '\st'
% 2. Please use '\textcolor{blue}{...}' if you would like to add some words
% 3. Please check that anything is wrong about your personal information
%%%--------------------------------

% make the title area
\maketitle

% As a general rule, do not put math, special symbols or citations
% in the abstract or keywords.
\begin{abstract}

We propose a Video Colorization with Hybrid Generative Adversarial Network (VCGAN), an improved approach to video colorization using end-to-end learning and recurrent architecture. The VCGAN addresses two prevalent issues in the video colorization domain: Temporal consistency and the unification of colorization network and refinement network into a single architecture. To enhance colorization quality and spatiotemporal consistency, the mainstream of the generator in VCGAN is assisted by two additional networks, \emph{i.e.,} global feature extractor and placeholder feature extractor, respectively. The global feature extractor encodes the global semantics of grayscale input to enhance colorization quality, whereas the placeholder feature extractor serves as a feedback connection to encode the semantics of the previous colorized frame in order to maintain spatiotemporal consistency. If changing the input for placeholder feature extractor as grayscale input, the hybrid VCGAN also has the potential to colorize single images. To improve the color consistency of far frames, we propose a dense long-term loss that minimizes the temporal disparity of every two remote frames. Trained with colorization and temporal losses jointly, VCGAN strikes a good balance between video color vividness and spatiotemporal continuity. Experimental results demonstrate that VCGAN produces higher-quality and temporally more consistent colorful videos than existing approaches.

\end{abstract}

% Note that keywords are not normally used for peerreview papers.
\begin{IEEEkeywords}
Video Colorization, Generative Adversarial Networks, Placeholder Feature Extractor.
\end{IEEEkeywords}

% For peer review papers, you can put extra information on the cover
% page as needed:
% \ifCLASSOPTIONpeerreview
% \begin{center} \bfseries EDICS Category: 3-BBND \end{center}
% \fi
%
% For peerreview papers, this IEEEtran command inserts a page break and
% creates the second title. It will be ignored for other modes.
\IEEEpeerreviewmaketitle

\section{Introduction}
% The very first letter is a 2 line initial drop letter followed
% by the rest of the first word in caps.
% 
% form to use if the first word consists of a single letter:
% \IEEEPARstart{A}{demo} file is ....
% 
% form to use if you need the single drop letter followed by
% normal text (unknown if ever used by the IEEE):
% \IEEEPARstart{A}{}demo file is ....
% 
% Some journals put the first two words in caps:
% \IEEEPARstart{T}{his demo} file is ....
% 
% Here we have the typical use of a "T" for an initial drop letter
% and "HIS" in caps to complete the first word.
\IEEEPARstart{T}{HERE} are many legacy movies and historical videos in black-and-white format. Restricted by the photography technology at that time, it was extremely hard to preserve color information. If the grayscale videos are painted with reasonable colors, they could show the vividness of the past time. Recently, the convolutional neural networks (CNNs) automate the process of grayscale image colorization \cite{cheng2015deep, larsson2016learning, isola2017image, zhang2016colorful, iizuka2016let, deshpande2017learning, royer2017probabilistic, cao2017unsupervised, guadarrama2017pixcolor, zhao2019pixelated, vitoria2020chromagan, zhao2020scgan, su2020instance, isola2017image, lei2019fully, kouzouglidis2019automatic, thasarathan2019automatic, jampani2017video, zhang2019deep, wan2020automated}. To predict plausible colorized images, researchers combined many objective functions such as L1 loss, MSE loss, perceptual loss \cite{johnson2016perceptual}, KL loss \cite{deshpande2017learning}, and classification loss on each pixel \cite{zhang2016colorful} or advanced training schemes like adversarial training \cite{goodfellow2014generative} and coarse-to-fine scheme \cite{wang2018high}. However, those image colorization algorithms cannot be directly utilized to colorize grayscale videos since they are unable to learn spatiotemporal consistency. Since adjacent frames in a video are temporally correlated, the additional spatiotemporal constraints are significant for video colorization applications.

\begin{figure}[t]
\centering
\includegraphics[width=\linewidth]{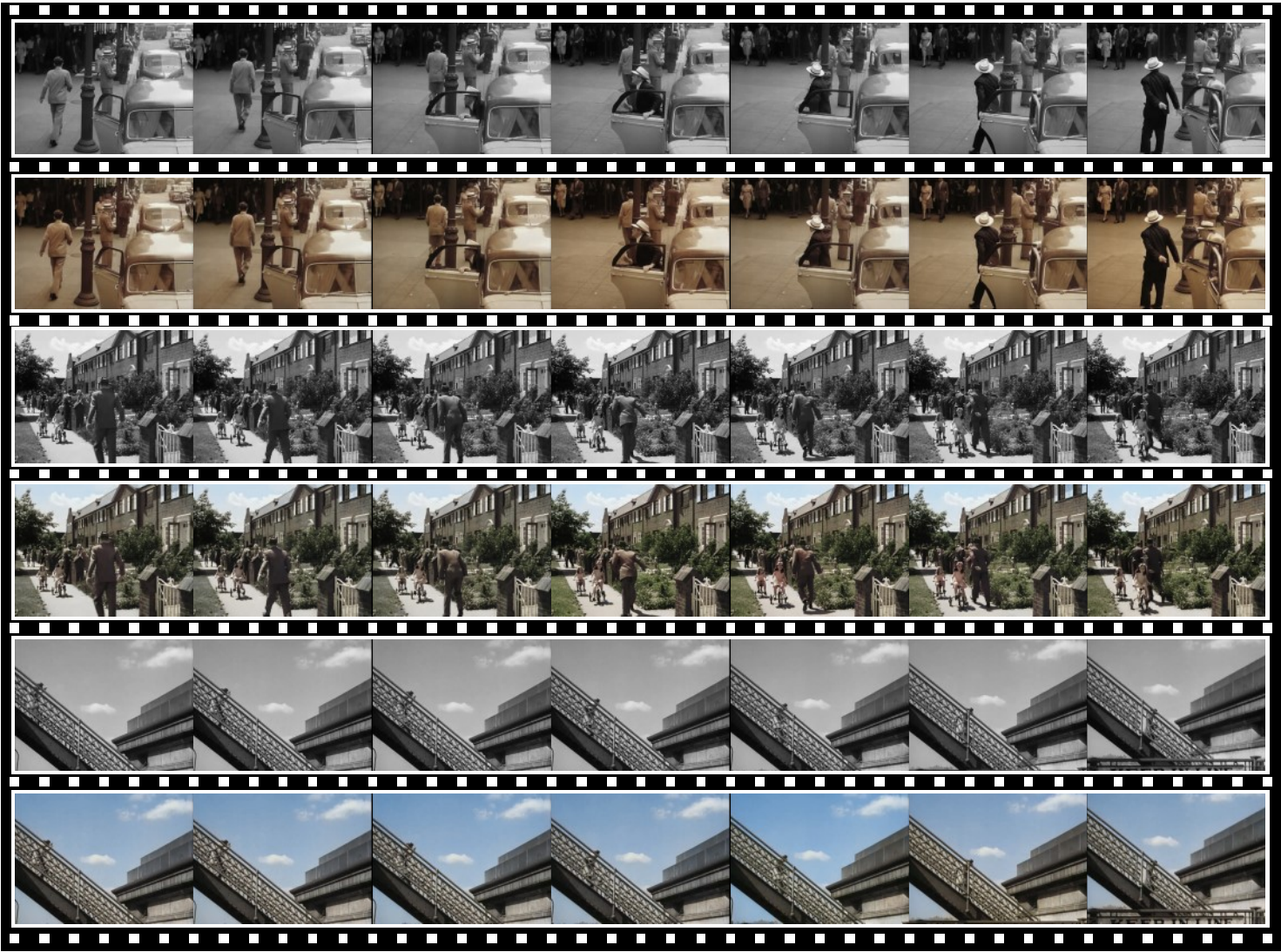}

\vspace{-2mm}

\caption{Colorization results of a 1948 American grayscale film ``The Naked City'' by the proposed VCGAN. Different rows represent different scenes in the film. The interval of frames equals to 5. Please see \textbf{https://github.com/zhaoyuzhi/VCGAN} for supplementary materials.}
\label{movie}

\vspace{-2mm}

\end{figure}

Existing video colorization methods can be categorized into three classes: exemplar-guided \cite{levin2004colorization, yatziv2006fast, sheng2013video, reinhard2001color, jampani2017video, zhang2019deep}, task-independent \cite{bonneel2015blind, lai2018}, and fully-automatic \cite{lei2019fully, kouzouglidis2019automatic, thasarathan2019automatic}. On one hand, earlier video colorization methods are often based on exemplars such as color scribbles and strokes \cite{levin2004colorization, yatziv2006fast, sheng2013video}. On the other hand, to alleviate the effort of selecting proper examples, task-independent video colorization methods \cite{bonneel2015blind, lai2018} post-process framewise colorization results by adding temporal coherence. For instance, Lai \emph{et al.} \cite{lai2018} utilized a temporal smoothing network to minimize the color differences between the two consecutive frames that are colorized individually using image colorization algorithms. However, the performance of these methods is limited by the image colorization algorithms. Furthermore, fully-automatic methods \cite{lei2019fully, kouzouglidis2019automatic, thasarathan2019automatic} learn the mapping from continuous grayscale frames to colorful frames. The mapping is normally implemented by a neural network, \emph{e.g.,} 3D-CNN \cite{kouzouglidis2019automatic}, two-step network \cite{lei2019fully}. On one hand, 3D-CNN requires a large memory footprint for a long sequence (\emph{e.g.,} each segment is independently colorized due to the GPU memory limit). On the other hand, two-step network misses the first frame at inference, in addition to requiring large memory footprints. Moreover, the balance between color vividness and video continuity becomes another issue.

In order to address these issues, we propose to combine both image and video colorization into a hybrid architecture VCGAN. There are three main benefits of the hybrid model: 1) One model can be applied to both image and video colorization; 2) It provides reference colorized frame as a prior for the colorization of the following grayscale frames, which avoids the requirement of temporal refinement network; 3) The proposed hybrid architecture strikes a good balance between color vividness and video continuity. Firstly, we assume the continuous frames satisfy the Markov chain and its transition function is implied in the model. Also, a video can include many same frames. Therefore, it is possible to combine both image and video colorization into the same architecture. Secondly, since the VCGAN has the ability to process a single image (\emph{i.e.,} the first frame of video), there are no frames lost compared with \cite{lei2019fully}. Thirdly, we propose a two-stage training schedule and use the adversarial training \cite{goodfellow2014generative} with a dense long-term loss. They help VCGAN achieve good colorization quality and maintain spatiotemporal continuity.

The proposed VCGAN generator architecture includes a mainstream encoder, a mainstream decoder, and two feature extractors. The first feature extractor extracts the semantics of the input grayscale frame, which provides high-level information for the network to better learn colors for objects \cite{larsson2016learning, zhang2016colorful, iizuka2016let, zhang2017real}. The second feature extractor makes VCGAN relate every two neighbouring frames for video colorization. It enhances the spatiotemporal continuity of output frames. It uses the same architecture as global feature extractor but receives the last colorized frame for video colorization. If changing the input as grayscale frame of current time, the VCGAN becomes an image colorization model. The output features of both extractors are concatenated to the mainstream encoder, which are then jointly fed into decoder. Therefore, the VCGAN generator can utilize these information to learn a good video colorization. In addition, we adopt a patch-based discriminator for adversarial learning.
% In addition, we adopt a patch-based structure for the VCGAN discriminator.
% The weights of two feature extractors are loaded from ResNet-50 \cite{he2016deep}.

Regarding the optimization, we define a two-stage training schedule for VCGAN including single image and video colorization, respectively. Since the total numbers and the diversity of frames in the video datasets are much less than image datasets, we first use the large-scale image dataset ImageNet \cite{russakovsky2015imagenet} to train VCGAN. The first stage provides good initialization weights, which ensures VCGAN has plausible image colorization quality. Then, at the second stage, it is optimized with both colorization and spatiotemporal smoothing objectives using video datasets such as DAVIS \cite{perazzi2016benchmark} and Videvo \cite{Videvo}. In addition, we improve the temporal smoothness of colorized frames by enforcing an additional dense long-term loss at the second stage. It models the dense connections of every remote frame, which is beneficial for VCGAN to maintain the color continuity for distant frames. The adversarial training is used to enhance the color vividness. We evaluate the proposed VCGAN in terms of both image and video colorization quality on the benchmark datasets. Experimental results demonstrate that VCGAN can produce high-quality colorizations than the well-known methods. Some results produced by VCGAN are shown in Figure \ref{movie}.

%Regarding the implementation, there are two training stages in VCGAN including single frame and video processing. The first stage provides good initialization for VCGAN and then it is optimized with colorization and spatiotemporal smoothing objectives jointly at second stage.

In general, there are three main contributions of this paper:

1) A hybrid recurrent VCGAN framework is proposed to integrate both image and video colorization applications;

2) A dense long-term loss is proposed to minimize the flicking artifacts of generated frames;

3) Comprehensive experiments are conducted to evaluate the VCGAN architecture on both single image and video colorization applications. The VCGAN achieves state-of-the-art performances on benchmark datasets compared with some well-known algorithms.

\begin{figure*}[t]
\centering
\includegraphics[width=\linewidth]{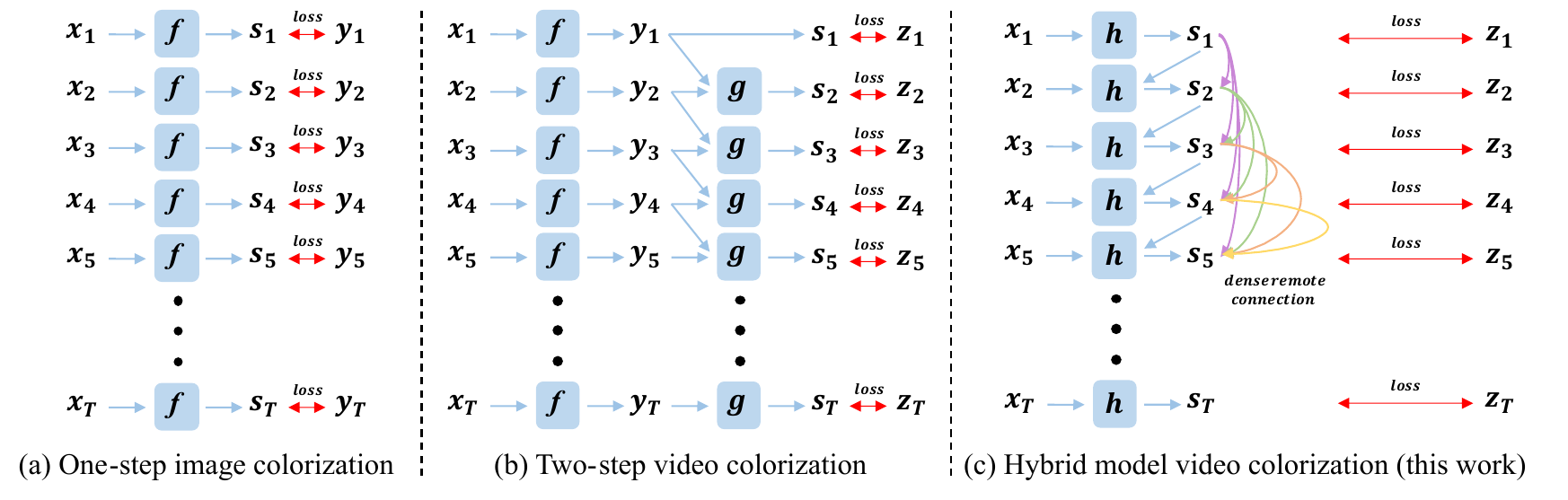}

\vspace{-4mm}

\caption{Illustration of different connection types of (a) One-step image colorization \cite{isola2017image, cheng2015deep, larsson2016learning, zhang2016colorful, iizuka2016let}, (b) Two-step video colorization \cite{lei2019fully}, and (c) the proposed hybrid VCGAN, where $x$ is the input, and $s$ is the video colorization result. $f$, $g$, and $h$ represent CNNs. The color lines in (c) indicate that the dense remote connections of generated frames modeled by VCGAN. The losses are computed between $s$ and ground truth $y$ (image colorization) or $z$ (video colorization).}

\vspace{-0mm}

\label{define}
\end{figure*}

\vspace{-0mm}

%-------------------------------------------------------------------------
\section{Related Work}

\textbf{Image Colorization.} There were two categories of image colorization methods: exemplar-based and fully-automatic. The exemplar-based methods are based on additional user-given information such as color scribbles \cite{levin2004colorization, yatziv2006fast, sheng2013video, zhang2017real} and example colorful images \cite{reinhard2001color, welsh2002transferring, tai2005local, liu2008intrinsic}. For instance, Levin \emph{et al.} \cite{levin2004colorization} assumed adjacent pixels with the same illuminances should have similar colors and developed an optimization-based system based on the assumption. Welsh \emph{et al.} \cite{welsh2002transferring} attached colors from example images to grayscale input by matching spatial features of them. However, these algorithms require accurate hints (\emph{e.g.,} color pixels or similar RGB images) for producing high-quality colorizations, which is labor-intensive.

To alleviate the effort of selecting proper references, fully-automatic image colorization methods \cite{cheng2015deep, larsson2016learning, isola2017image, zhang2016colorful, iizuka2016let, deshpande2017learning, royer2017probabilistic, cao2017unsupervised, guadarrama2017pixcolor, zhao2019pixelated, vitoria2020chromagan, zhao2020scgan, su2020instance} directly learn the mapping from grayscale images to their color embeddings based on deep learning. Cheng \emph{et al.} \cite{cheng2015deep} firstly utilized a deep neural network to colorize images based on three levels of features. However, the performance is limited due to hand-crafted features and a tiny network structure. To improve generation quality, researchers used semantics extracted by pre-trained VGG-Net \cite{simonyan2014very} or ResNet \cite{he2016deep}. For instance, Larsson \emph{et al.} \cite{larsson2016learning} adopted a VGG-Net-based hyper-column to extract multi-level representation of grayscale. Iizuka \emph{et al.} \cite{iizuka2016let} used two-stream networks for extracting both low-level and high-level information. While Zhang \emph{et al.} \cite{zhang2016colorful} directly adopted VGG-16 as backbone with a color classification loss and category-balancing technique. To augment the colorization for significant objects in an image, Zhao \emph{et al.} \cite{zhao2020scgan} used saliency map to aid the learning of colorization and Su \emph{et al.} \cite{su2020instance} includes instance segmentation in colorization system.

%In order to further refine visual artifacts for those methods, some regularization terms such as gradients \cite{deshpande2017learning} and segmentations \cite{zhao2018pixel} were proposed. More recently, Vitoria \emph{et al.} \cite{vitoria2020chromagan} simplified the structure of \cite{iizuka2016let} and added adversarial loss \cite{goodfellow2014generative}. Su \emph{et al.} \cite{su2020instance} combined instance and entire-image colorization into one architecture. Thus, it focuses more on detected significant instances, which leads to better colorization reality.

\textbf{Video Colorization.} There are three classes of video colorization algorithms: exemplar-guided \cite{reinhard2001color, levin2004colorization, yatziv2006fast, sheng2013video, jampani2017video, zhang2019deep, wan2020automated}, task-independent \cite{bonneel2015blind, lai2018} and fully-automatic \cite{lei2019fully, kouzouglidis2019automatic, thasarathan2019automatic}. The earlier works were mainly exemplar-guided including propagating the user scribbles \cite{levin2004colorization, yatziv2006fast, sheng2013video}, attaching the colors from colorized frames \cite{jampani2017video} or given images \cite{reinhard2001color} to the rest of frames. Recently, CNNs improve colorization quality since it effectively extracts features from the input \cite{larsson2016learning, zhang2016colorful, iizuka2016let}. For instance, Zhang \emph{et al.} \cite{zhang2019deep} matched the features between the reference image and input frames to guide colorization. Jampani \emph{et al.} \cite{jampani2017video} used few colorized frames as references and then propagated them to the whole video. However, their results are plausible when the scene disparity of examples and grayscale frames can be ignored.

Many image colorization algorithms obtain good colorization quality; however, directing using them to each video frame independently often leads to temporal inconsistencies. Thus, the task-independent methods were proposed to explicitly encode the temporal consistency of the independently colorized frames. Bonneel \emph{et al.} \cite{bonneel2015blind} addressed the issue by minimizing the disparity of warped frame and next frame with least-square energy. Lai \emph{et al.} \cite{lai2018} introduced a transformation network that post-processes the frames, with an optical flow guidance. The network was trained by both temporal and perceptual loss \cite{johnson2016perceptual} to strike a balance between temporal coherence and spatial quality. However, the refined frames are still not continuous enough, since the image colorization and temporal refinement networks are not trained collaboratively. To further automate the video colorization pipeline, Lei \emph{et al.} \cite{lei2019fully} proposed a multimodal automatic system that produced four possible colorized videos. To enhance the color consistency, they performed the K-nearest neighbor (KNN) search that builds a connection between color and spatial location. However, the generated images are not colorful enough.

\textbf{Generative Adversarial Network for Colorization.} GAN was first proposed by Goodfellow \emph{et al.} \cite{goodfellow2014generative}, including two neural networks (\emph{i.e.,} generator and discriminator) that compete against each other. For colorization, GAN was used to enhance the vividness of colorized images \cite{isola2017image} or produce diverse results \cite{cao2017unsupervised, zhu2017toward}. Isola \emph{et al.} \cite{isola2017image} proposed a general Pix2Pix framework for paired images transformation. Experimental analysis proved that adversarial training strategy helps in preserving details and enhancing the perceptual quality. It was enhanced by Pix2PixHD framework \cite{wang2018high} for high-resolution images. To obtain diverse colorization, Cao \emph{et al.} \cite{cao2017unsupervised} directly added noise to the first three layers of encoder while Zhu \emph{et al.} \cite{zhu2017toward} introduced a cLR-GAN model including variational training to strengthen the output diversity.

\begin{figure*}[t]
\begin{center}
\centering
\includegraphics[width=\linewidth]{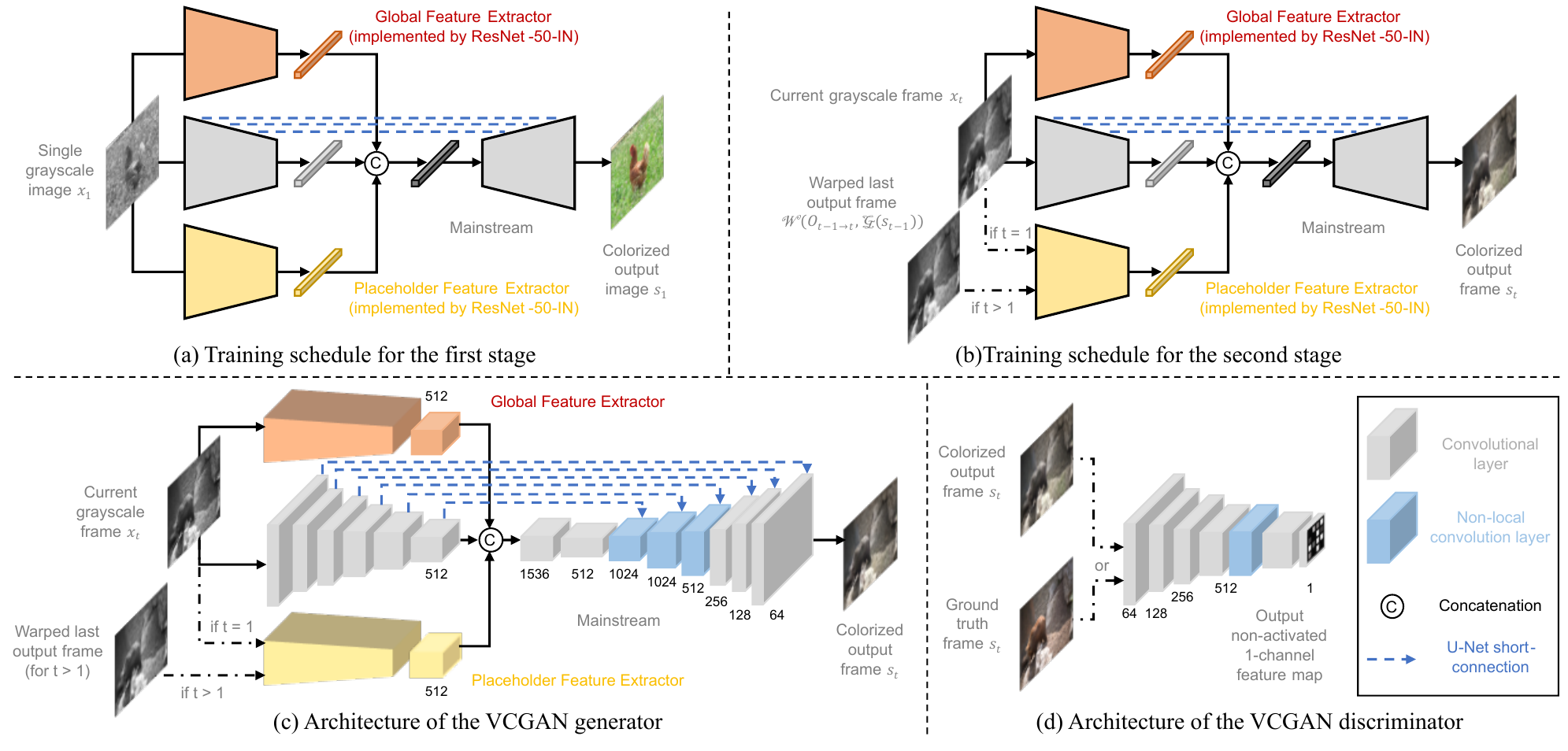}
%\caption{Illustration of the architecture of the proposed VCGAN. It consists of the generator (left part) and the discriminator (middle part). The right part annotates the different blocks. The generator receives a grayscale frame input at the first frame position. For the following frames, the generator receives a grayscale frame of the current time and last colorized frame as input. The discriminator receives the colorized frame or ground truth frame.}

\vspace{-3mm}

\caption{Illustration of the training schedules for (a) the first stage and (b) the second stage. Illustration of the architectures of (c) the VCGAN generator and (d) the VCGAN discriminator, where the numbers of channels and notations of layers/blocks/operations are attached. More detailed architectures can be found in supplementary materials (\textbf{https://github.com/zhaoyuzhi/VCGAN}).}

\vspace{-5mm}

\label{net}
\end{center}
\end{figure*}

%\begin{figure*}[t]
%\begin{center}
%\centering
%\includegraphics[width=17.8cm]{pdf/subnet.pdf}
%\caption{Architecture of the proposed VCGAN discriminator and detailed structure of GLE Block in generator. The patch-based discriminator transforms input image to a non-activated feature map. The GLE Block is consisted of two sequential RDC Block and one RDLC Block, with short connection between each of them.}
%\label{subnet}
%\end{center}
%\end{figure*}

\vspace{-0mm}

\section{Methodology}

\subsection{Problem Formulation}

Given a grayscale input video, the output colorized video should satisfy two conditions. Firstly, the color of generated frames should be similar to ground truth. Secondly, the temporal disparity of adjacent frames in the colorized video should be small, \emph{i.e.,} there is almost no flickering effect in the colorized video. Both of the conditions are equally crucial for video colorization.

Suppose the frames of input grayscale video with length $T$ is represented as a sequence $X=\left\{x_1,x_2,...,x_T\right\}$. The corresponding results processed by image colorization algorithms can be represented as $Y=\left\{y_1,y_2,...,y_T\right\}$ and the ground truth colorful video frames are $Z=\left\{z_1,z_2,...,z_T\right\}$. Note that, the framewise color similarity of $Y$ should be highly comparable with $Z$. However, the frames of $Y$ are temporally discontinuous, since the single image colorization methods \cite{isola2017image, cheng2015deep, larsson2016learning, zhang2016colorful, iizuka2016let} only learn one-step conditional distribution $p(Y | X)$. To address the issue of discontinuity, current video colorization methods \cite{lei2019fully, lai2018, zhang2019deep} finetune the results from $Y$ by another refinement network. They learn a two-step joint distribution $p(Z | Y)p(Y | X)$. Under these conditions, the mapping function can be factorized as:
\begin{equation}
p(Z | X,Y)=\prod_{t=2}^{T}p(z_t | y_t,y_{t-1})p(y_t | x_t)p(y_{t-1} | x_{t-1}).
\label{equ1}
\end{equation}

Specifically, the generated frame contains the information of the previous frame $x_{t-1}$ and the current frame $x_t$; however, there is no direct connection between them. Normally, the $p(y_t | x_t)$ is implemented by an image colorization network, which is trained to generate inconsistent $Y$ by individually colorizing grayscale frames. Then, a refinement network is used to post-process continuous two frames, \emph{i.e.,} $p(z_t | y_t,y_{t-1})$. It is difficult to control the video consistency of generated frames if the networks are trained individually \cite{lai2018}. Although adopting a joint training scheme \cite{lei2019fully}, the system is too large thus the optimization is difficult to perform. To address this problem, we alternatively learn the direct mapping from $X$ to $Z$. Therefore, the learning process of the proposed VCGAN is represented as:
\begin{equation}
p(Z | X)=\prod_{t=2}^{T}p(z_t | x_t,z_{t-1})p(z_1 | x_1).
\label{equ2}
\end{equation}

The two conditional distributions $p(z_t | x_t,z_{t-1})$ and $p(z_1 | x_1)$ are combined into one model by a placeholder feature extractor. In addition, the proposed VCGAN is recurrent since the previous colorized frame becomes the input for the colorization process of the next frame, which is optimized to be close to ground truth $z_{t-1}$. Thus, VCGAN is a hybrid end-to-end model that colorizes grayscale frames sequentially. Since there is no previous frame as a guidance, VCGAN generates the first frame $z_1$ only based on the initial input frame $x_1$. For such case, it is viewed as a special video colorization issue, \emph{i.e.,} only one frame in the video. For the following frames, VCGAN does not produce intermediate variables and it is optimized by colorization and temporal smoothing objectives jointly. The recurrent architecture explicitly enforces VCGAN to synthesize more temporally consistent results. Figure \ref{define} illustrates the different connection types of the two aforementioned representative video colorization approaches and VCGAN.

\vspace{-0mm}

%-------------------------------------------------------------------------

\subsection{Two-stage Training Schedule}

In order to ensure that VCGAN produces perceptually plausible colorizations, the training process is divided into two stages. At the first training stage, the VCGAN performs single image colorization, as shown in Figure \ref{net} (a). The large ImageNet dataset \cite{russakovsky2015imagenet} is utilized for training since it contains much more diverse modes and categories compared with common video datasets \cite{perazzi2016benchmark, Videvo}. After its convergence, VCGAN is eligible to produce a single colorful image with high pixel accuracy.

At the second training stage, VCGAN is trained as a Markov Chain that performs a sliding window scheme to select continuous frames. For the first frame colorization, the VCGAN learns $p(z_1 | x_1)$ (see Equation (\ref{equ2})). For the following frames (\emph{e.g.,} time $t$), we consider the relations between every two neighbouring frames, \emph{i.e.,} VCGAN learns $p(z_t | x_t, z_{t-1})$ (see equation (\ref{equ2})). The output of time $t$-1 is first converted to grayscale and warped using forward flow from time $t$-1 to $t$. Then, the warped image replaces the grayscale input at time $t$ for the placeholder feature extractor, as shown in Figure \ref{net} (b). The processes can be defined as:
%For the first frame, the inputs for two feature extractors and mainstream are the same, \emph{i.e.,} VCGAN learns $p(z_1 | x_1)$ (see equation (\ref{equ2})). For the following frames (\emph{e.g.,} time $n$), the output of time $n-1$ is first converted to grayscale and warped using forward flow from time $n-1$ to $n$. Then, the warped image replaces the grayscale input for the placeholder feature extractor. The processes can be defined as:
\begin{equation}
s_{t} = \left\{
\begin{aligned}
G(x_{1}) & , & t = 1, \\
G(x_{t}, i_{t}) & , & t > 1.
\end{aligned}
\right.
\label{two_stage_training}
\end{equation}
\begin{equation}
i_t = \mathscr{W} (O_{t-1 \to t}, \mathscr{G} ( p_{t-1} )) ,
\label{two_stage_training2}
\end{equation}
where $s_{t}$ and $i_{t}$ are the output of VCGAN generator and input for placeholder feature extractor when $t>$ 1. The network $G(*)$ represents the VCGAN generator. The operator $\mathscr{W}(*)$ warps input frame under the guidance of given optical flow $O_{t-1 \to t}$, and the operator $\mathscr{G}(*)$ converts RGB images to grayscale by a linear transformation. Note that $\mathscr{W}(*)$ and $\mathscr{G}(*)$ are fixed; therefore $i_t$ is proportional to $s_{t-1}$. Therefore, VCGAN can utilize the information from last output, which satisfies the representation of equation (\ref{equ2}).

This design unifies both image and video colorization. Compared with single image colorization algorithms \cite{larsson2016learning, zhang2016colorful, iizuka2016let}, the placeholder feature extractor reserves a place for recurrent feedback. Moreover, it encourages VCGAN to minimize the color discrepancy between neighbouring frames.

\subsection{VCGAN Architecture}

The hierarchical VCGAN generator consists of three main parts: global feature extractor, placeholder feature extractor, and mainstream encoder-decoder, as shown in Figure \ref{net} (c). The mainstream adopts U-Net structure \cite{ronneberger2015u} that executes skip connection between each encoder layer $i$ and decoder layer $n$-$i$ with the same spatial resolution, where $n$ is the total number of mainstream layers. It promotes the decoder to preserve low-level details and facilitates the convergence of the whole system since the gradients easily pass to encoder layers. The non-local blocks \cite{zhang2018self} are attached to bottom layers of the decoder, which strengthen the details using cues from spatially related pixels.

The global feature extractor and placeholder feature extractor utilize a fully convolutional ResNet-50-IN network \cite{he2016deep} architecture, both of which are pre-trained on ImageNet \cite{russakovsky2015imagenet}. Since the colorization highly depends on global information \cite{zhang2016colorful, iizuka2016let, larsson2016learning}, the global feature extractor distills semantics from input effectively. While the placeholder feature extractor reserves the information of the last frame to enhance temporal consistency. The outputs of the two feature extractors are concatenated to mainstream encoder for feature fusion.

We adopt the PatchGAN discriminator \cite{isola2017image} to produce a 1-channel matrix corresponding to input resolutions, as shown in Figure \ref{net} (d). It contains fewer parameters than the original 1$\times$1 PixelGAN yet enhances the perceptual quality of generated samples. It also encourages sharper edges and colors.

\subsection{Loss Functions}

%Directly optimizing a conditional GAN framework with adversarial loss often leads to failure. Thus, we pre-train the generator to produce relatively plausible results to stabilize GAN training based on the following objective:
At the first stage, VCGAN is learned to produce accurate image colorization. The loss function of the first stage is:
\begin{equation}
L_{1st} = \lambda_1 L_1 + \lambda_p L_p ,
\label{1st}
\end{equation}
where $L_1$ and $L_p$ denote pixel-level reconstruction loss and perceptual loss \cite{johnson2016perceptual}, respectively. $\lambda_1$ and $\lambda_p$ are trade-off coefficients. Specifically, the losses are defined as:
\begin{equation}
L_1 = \mathbb{E}[||s_t - z||_1] ,
\label{1}
\end{equation}
\begin{equation}
L_p = \mathbb{E}[||\phi_l (s_t) - \phi_l (z)||_1] ,
\label{p}
\end{equation}
where $s_t$ (see Equation (\ref{two_stage_training})) and $z$ represent the colorized image and corresponding ground truth, respectively. At the first stage, $t$=1. $\phi_l (*)$ produces the features from the $l$-th layer of a pre-trained network. In our experiment, the $conv_{4\_3}$ layer of VGG-16 network \cite{simonyan2014very} is adopted.

At the second stage, we train VCGAN generator and discriminator alternatively and include optical flow for matching spatial location. The overall loss function is defined as:
\begin{equation}
L_{2nd} = \lambda_1 L_1 + \lambda_p L_p + \lambda_G L_G + \lambda_{st} L_{st}  + \lambda_{dlt} L_{dlt} ,
\label{2nd}
\end{equation}
where $L_G$, $L_{st}$, and $L_{dlt}$ indicate GAN loss, short-term loss, and dense long-term loss, respectively. $\lambda_*$ are trade-off coefficients for each loss term.

We use the WGAN critic \cite{arjovsky2017wasserstein} and spectral normalization \cite{miyato2018spectral} in the adversarial training, which is defined as:
\begin{equation}
L_G = - \mathbb{E}[D(s_t)] ,
\label{g}
\end{equation}
\begin{equation}
L_D = \mathbb{E}[D(s_t)] - \mathbb{E}[D(z)] ,
\label{d}
\end{equation}
where Equations (\ref{g}) and (\ref{d}) constitute WGAN loss for generator $G(*)$ and discriminator $D(*)$, respectively. Due to spectral normalization attached to each convolutional layer of discriminator, VCGAN satisfies the 1-Lipschitz continuity.

%$\mathcal{P}$ $\mathfrak{W}$ $\mathcal{FPGD}$ $\mathscr{W}$  $\boldsymbol{GDs}$

To enforce temporal consistency, VCGAN should also learn connections for continuously generated frames. Suppose that there are $N$ continuous frames used for training in each iteration, the optical flow-based objectives include short-term loss and dense long-term loss are defined as:
\begin{equation}
%L_{st} = \mathbb{E}[ \sum_{n=2}^N M_{n-1 \to n} || G(x_n) - \mathscr{W} (O_{n-1 \to n}, G(x_{n-1})) ||_1] ,
L_{st} = \mathbb{E}[ \sum_{t=2}^N M_{t-1 \to t} || s_t - \mathscr{W} (O_{t-1 \to t}, s_{t-1}) ||_1] ,
\label{st}
\end{equation}
\begin{equation}
%L_{dlt} = \mathbb{E}[ \sum_{n=3}^N \sum_{m=1}^{n-2} M_{m \to n} || G(x_n) - \mathscr{W} (O_{m \to n}, G(x_{m})) ||_1] ,
L_{dlt} = \mathbb{E}[ \sum_{t=3}^N \sum_{m=1}^{t-2} M_{m \to t} || s_t - \mathscr{W} (O_{m \to n}, s_m) ||_1] ,
\label{lt}
\end{equation}
where $N$ is the numbers of frames in a batch, $s_m$ and $s_t$ are the colorized frames at time $m$ and $t$, respectively. $M_{m \to t}$ and $O_{m \to t}$ represent the non-occlusion mask \cite{lai2018} and real forward flow of colorful images between time $m$ and $t$, respectively. The operator $\mathscr{W}(*)$ warps input frame under the guidance of flow $O_{m \to t}$. By matching the pixel-wise non-occlusion region of the warped frame and current output, it enforces the temporal consistency of correctly warped regions. The short-term loss learns the color similarity for neighbouring frames. The dense long-term loss models each remote connection between two generated frames. Moreover, we follow the protocol in \cite{lai2018} to estimate mask:
\begin{equation}
M_{m \to t} = exp(- \alpha ||x_t, \mathscr{W} (O_{m \to t}, x_{m})||_2^2 ),
\label{mask}
\end{equation}
where the mask $M_{m \to t}$ indicating the non-occlusion regions of the warped image. The scale factor $\alpha$ enlarges the numerical disparity between occlusion and non-occlusion regions.

%The $N$ indicates the maximum frames processed in one iteration.

%------------------------------------------------------------------------
\begin{figure*}[h]
\centering
\includegraphics[width=\linewidth]{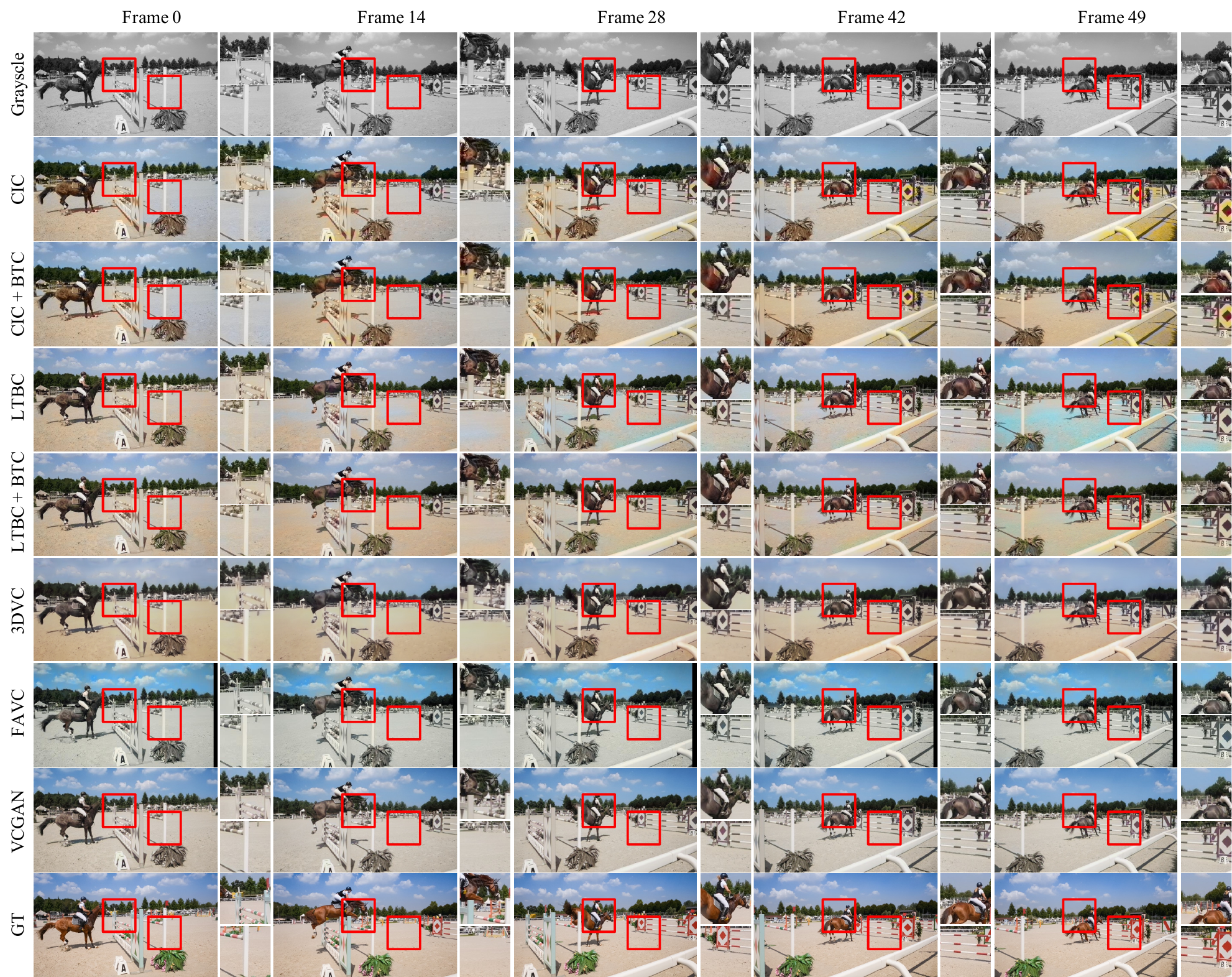}

\vspace{-2mm}

\caption{Colorization comparison on ``horsejump-high'' from DAVIS \cite{perazzi2016benchmark} dataset. The first and last rows include the grayscale and colorful ground truth frames, respectively. The middle rows include colorized results from state-of-the-art methods CIC \cite{zhang2016colorful}, CIC + BTC \cite{zhang2016colorful, lai2018}, LTBC \cite{iizuka2016let}, LTBC + BTC \cite{iizuka2016let, lai2018}, 3DVC \cite{kouzouglidis2019automatic}, FAVC \cite{lei2019fully}, and the proposed VCGAN. The red rectangles highlight inconsistent regions or strange colors for the baselines. Please refer to supplementary materials for full-range results (more frames generated by different methods and representative video clips).}
% Please refer to supplementary materials for more results.

\vspace{-5mm}

\label{compare1}
\end{figure*}

\begin{figure*}[h]
\centering
\includegraphics[width=\linewidth]{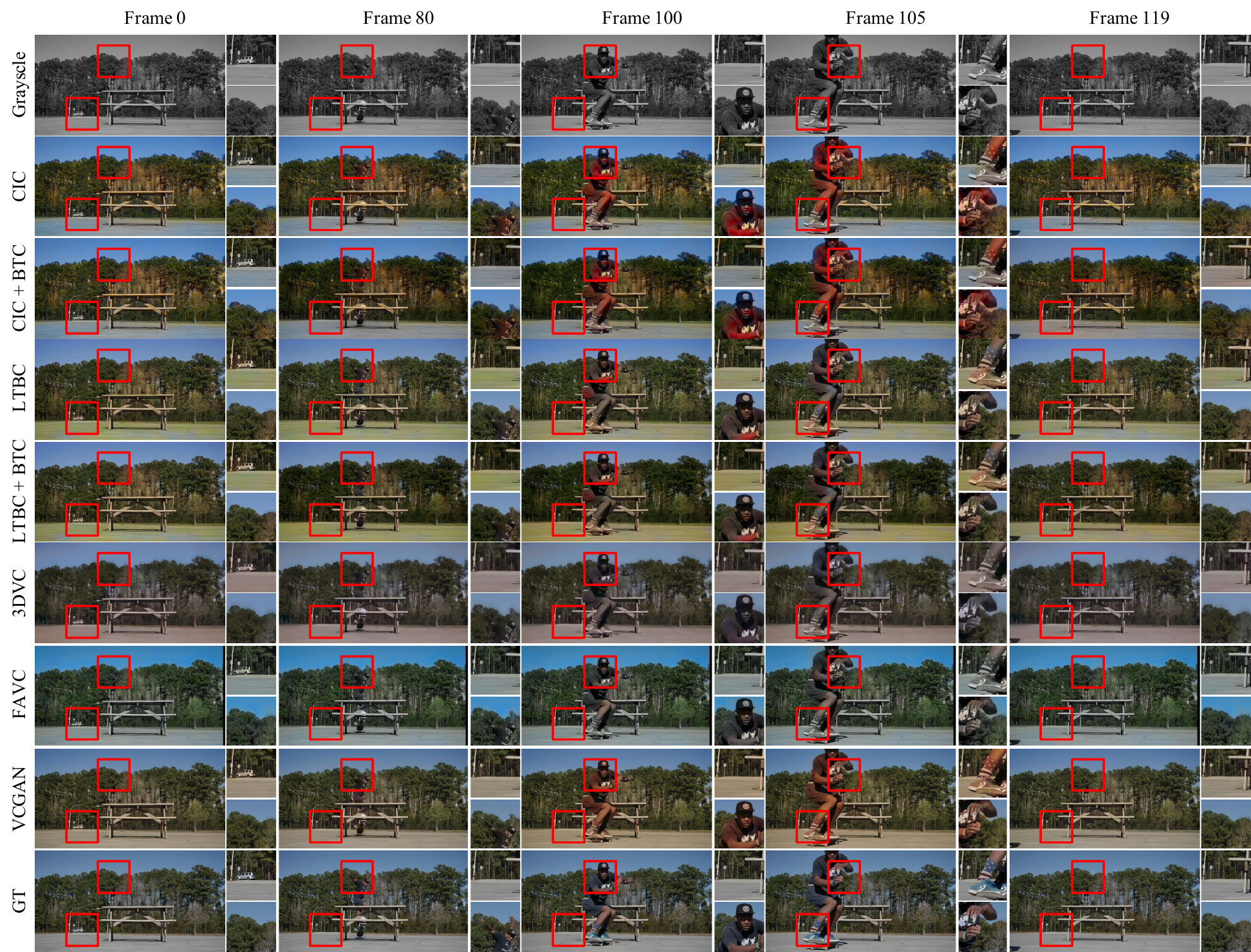}

\vspace{-2mm}

\caption{Colorization comparison on ``SkateboarderTableJump'' from Videvo \cite{Videvo} dataset. The first and last rows include the grayscale and colorful ground truth frames, respectively. The middle rows include colorized results from state-of-the-art methods CIC \cite{zhang2016colorful}, CIC + BTC \cite{zhang2016colorful, lai2018}, LTBC \cite{iizuka2016let}, LTBC + BTC \cite{iizuka2016let, lai2018}, 3DVC \cite{kouzouglidis2019automatic}, FAVC \cite{lei2019fully}, and the proposed VCGAN. The red rectangles highlight inconsistent regions or strange colors for the baselines.}

\vspace{-5mm}

\label{compare2}
\end{figure*}

\section{Experiment}

\subsection{Implementation Details}

\textbf{Dataset.} We use the entire ImageNet \cite{russakovsky2015imagenet} dataset (1281167 images with 1000 categories) at the first training stage. The images are resized to 256$\times$256. The images encoded as grayscale are excluded. At the second training stage, we utilize the DAVIS \cite{perazzi2016benchmark} and Videvo \cite{Videvo} datasets that contain 156 short videos (overall 29620 images). We assume each short video is equally important when selecting data for training. All training images are normalized to the range of [-1, 1].

\textbf{Network.} Both the generator and discriminator adopt LeakyReLU \cite{maas2013rectifier} activation function. The instance normalization \cite{ulyanov2016instance} is attached to each convolutional layer of both encoder and discriminator except the first and the last layers. Note that, the pre-trained ResNet-50-IN \cite{he2016deep} also adopts LeakyReLU \cite{maas2013rectifier} activation function and instance normalization \cite{ulyanov2016instance}. Specifically, to maintain more information while performing the down-sampling operation, the pooling layers of the original ResNet-50-IN architecture \cite{he2016deep, ulyanov2016instance} are replaced by convolutional layers with a stride of 2. At the final part of the network, an additional convolutional layer is added to reduce the dimension from 2048 to 512. We train this ResNet-50-IN from scratch following the hyper-parameter settings of \cite{he2016deep} until the ImageNet validation accuracy is high enough and stable. Then, the weights are loaded to the two feature extractors of VCGAN, while the weights of other layers of VCGAN are initialized with Xavier method \cite{glorot2010understanding}.

\textbf{Optimization.} For the first stage, the generator of VCGAN is trained with Equation (\ref{1st}) for 20 epochs. The learning rate is initialized to 2$\times$10$^{-4}$, which is halved after 10 epochs. For the second stage, we load the weights from the first stage for VCGAN generator. Then, the whole VCGAN is trained with Equation (\ref{2nd}) on continuous frames with 256p resolution and 480p resolution, for 500 epochs and 500 epochs, respectively. The initial learning rates for both generator and discriminator equal to 5$\times$10$^{-5}$. For 480p resolution, the learning rate is halved every 100 epochs. For a single category, we randomly sample $N$=5 successive frames at one iteration. The scale factor $\alpha$ of the non-occlusion mask $M$ (see Equations (\ref{st}), (\ref{lt}), and (\ref{mask}) is set to 50. For the optimization, we use Adam optimizer \cite{kingma2014adam} with $\beta_1$=0.5 and $\beta_2$=0.999. The batch size equals to 16 and 4 for the two stages, respectively.

The coefficients $\lambda_1$, $\lambda_p$, $\lambda_G$, $\lambda_{st}$, $\lambda_{dlt}$ are empirically set to 10, 5, 1, 3, 5, respectively. At the first stage, the VCGAN is trained on 4 NVIDIA Titan Xp GPUs (12 Gb). At the second stage, the training processes on 256p resolution and 480p resolution are performed on 4 NVIDIA Titan Xp GPUs (12 Gb) and 4 NVIDIA Tesla V100 GPUs (32 Gb), respectively. We implement the VCGAN using the PyTorch 1.0.0 framework with Python 3.6. The whole training of VCGAN takes approximately 14 days, where 10, 1, and 3 days for the first stage, second stage on 256p and 480p, respectively.

% They occupy approximately 10 Gb and 27 Gb memories per card on 256p resolution and 480p resolution, respectively.

\begin{table*}[h]
\begin{center}
\caption{Comparison of video colorization methods \cite{zhang2016colorful, iizuka2016let, vitoria2020chromagan, zhao2020scgan, lai2018, kouzouglidis2019automatic, lei2019fully} and the proposed VCGAN on DAVIS and Videvo datasets. The \textcolor{red}{\textbf{red}}, \textcolor{blue}{\textbf{blue}}, and \textcolor{green}{green} colors represent the best, the second-best, and the third-best performances, respectively.}
\label{table_sota}

\vspace{-2mm}

\begin{tabular}{lcccccccccc}
\hline
\multirow{2}{*}{Method} & \multicolumn{3}{c}{DAVIS} & \multicolumn{3}{c}{Videvo} & \multicolumn{3}{c}{Network Architecture} \cr & PSNR & SSIM & Warp Error & PSNR & SSIM & Warp Error & Semantic Model & Flow Estimator & Hybrid Model \\
\hline
\hline
Grayscale & 23.77 & 0.9484 & / & 25.31 & 0.9570 & / & / & / & / \\
CIC \cite{zhang2016colorful} & 22.44 & 0.9003 & 0.06055 & 21.79 & 0.8989 & 0.03317 & \checkmark & / & / \\
LTBC \cite{iizuka2016let} & 23.89 & 0.9130 & 0.05901 & 24.64 & 0.9237 & 0.03285 & \checkmark & / & / \\
ChromaGAN \cite{vitoria2020chromagan} & 23.70 & 0.9377 & 0.06023 & 23.88 & 0.9354 & 0.03319 & \checkmark & / & / \\
SCGAN \cite{zhao2020scgan} & 23.19 & 0.8959 & 0.05918 & 23.29 & 0.8549 & 0.03301 & \checkmark & / & / \\
\hline
CIC + BTC \cite{lai2018} & 21.48 & 0.8898 & 0.05170 & 21.02 & 0.8800 & 0.02891 & \checkmark & FlowNet2 & / \\
LTBC + BTC \cite{lai2018} & 22.45 & 0.9006 & 0.05144 & 22.81 & 0.9072 & 0.02995 & \checkmark & FlowNet2 & / \\
ChromaGAN + BTC \cite{lai2018} & 19.88 & 0.8896 & \textcolor{green}{\textbf{0.04955}} & 16.63 & 0.8289 & 0.02753 & \checkmark & FlowNet2 & / \\
SCGAN + BTC \cite{lai2018} & 19.35 & 0.8716 & \textcolor{blue}{\textbf{0.04902}} & 16.28 & 0.7455 & \textcolor{green}{\textbf{0.02715}} & \checkmark & FlowNet2 & / \\
3DVC \cite{kouzouglidis2019automatic} & \textcolor{blue}{\textbf{23.43}} & \textcolor{blue}{\textbf{0.9115}} & 0.05125 & \textcolor{blue}{\textbf{24.28}} & \textcolor{blue}{\textbf{0.9200}} & \textcolor{blue}{\textbf{0.02659}} & / & 3D Conv & / \\
FAVC \cite{lei2019fully} & \textcolor{green}{\textbf{22.98}} & \textcolor{green}{\textbf{0.9055}} & 0.06002 & \textcolor{green}{\textbf{23.47}} & \textcolor{green}{\textbf{0.9183}} & 0.03236 & \checkmark & PWC-Net & / \\
\hline
VCGAN & \textcolor{red}{\textbf{23.77}} & \textcolor{red}{\textbf{0.9196}} & \textcolor{red}{\textbf{0.04871}} & \textcolor{red}{\textbf{25.11}} & \textcolor{red}{\textbf{0.9264}} & \textcolor{red}{\textbf{0.02502}} & \checkmark & PWC-Net & \checkmark \\
\hline
\end{tabular}
\end{center}

\vspace{-3mm}

\end{table*}

%\begin{table*}[h]
%\begin{center}
%\caption{Comparison of state-of-the-art video colorization methods \cite{zhang2016colorful, iizuka2016let, lai2018, lei2019fully} and proposed VCGAN on 480p videos.}
%\begin{tabular}{lccccccccccc}
%\hline
%\multirow{2}{*}{Method} & \multicolumn{4}{c}{DAVIS} & \multicolumn{4}{c}{Videvo} & \multicolumn{2}{c}{Network Architecture} \cr & LPIPS & PSNR & SSIM & Warp Error & LPIPS & PSNR & SSIM & Warp Error & Estimator & Hybrid Model \\
%\hline
%Grayscale & 0.2205 & 23.77 & 0.9484 & / & 0.2184 & 25.31 & 0.9570 & / & / & / \\
%LTBC \cite{iizuka2016let} & 0.1900 & 23.89 & 0.9130 & 0.05901 & 0.1951 & 24.64 & 0.9237 & 0.03285 & / & / \\
%CIC \cite{zhang2016colorful} & 0.2189 & 22.44 & 0.9003 & 0.06055 & 0.2456 & 21.79 & 0.8989 & 0.03317 & / & / \\
%\hline
%\hline
%LTBC + BTC \cite{iizuka2016let, lai2018} & \textbf{0.2208} & 22.45 & 0.9006 & 0.05144 & \textbf{0.2043} & 22.81 & \textbf{0.9072} & 0.02995 & FlowNet2 & / \\
%CIC + BTC \cite{zhang2016colorful, lai2018} & 0.2464 & 21.48 & 0.8898 & 0.05170 & 0.2521 & 21.02 & 0.8800 & 0.02891 & FlowNet2 & / \\
%FAVC \cite{lei2019fully} & 0.3255 & 16.89 & 0.4531 & 0.06002 & 0.3186 & 17.91 & 0.5600 & 0.03236 & PWC-Net & / \\
%VCGAN & 0.2368 & \textbf{23.19} & \textbf{0.9030} & \textbf{0.05031} & 0.2527 & \textbf{22.97} & 0.9066 & \textbf{0.02702} & PWC-Net & \checkmark \\
%\hline
%\end{tabular}
%\end{center}
%\label{table_sota}
%\end{table*}

\subsection{Experiment Settings}

\textbf{Dataset.} Following \cite{lei2019fully, lai2018}, we perform the evaluations on DAVIS \cite{perazzi2016benchmark} and Videvo \cite{Videvo} testing set. The DAVIS dataset includes 30 short videos, each of which contains approximately 100 frames. The Videvo dataset consists of 20 videos and there are about 300 frames in each clip. Although different approaches may produce images of diverse resolutions, all the result images are generated and resized to match the image resolution of ground truth for fairness. Moreover, since the proposed VCGAN can generate a single colorful image using weights of the first stage, we assess its colorization quality by colorizing single images. We use the 10000 ImageNet validation images \cite{russakovsky2015imagenet} as same as \cite{zhang2016colorful, larsson2016learning, vitoria2020chromagan, zhao2020scgan}.

\textbf{PSNR and SSIM \cite{wang2004image}.} To represent the fidelity of generated image, we apply PSNR to calculate the pixel-level error. Since PSNR is not highly relevant to the human visual system, we also adopt SSIM \cite{wang2004image} to estimate the structural similarity (especially luminance, contrast, structure).

\textbf{Top-5 Accuracy.} To estimate the semantic interpretability, we adopt the Top-5 Accuracy. It is only for evaluating image colorization quality based on a pre-trained VGG-16 network.

% \cite{simonyan2014very}

%\noindent \textbf{LPIPS \cite{zhang2018unreasonable}.} Recently, the deep layers of a trained classification network have been shown advance in representing the semantic features of input images. To represent the perceptual distance of two images, Zhang \emph{et al.} \cite{zhang2018unreasonable} proposed a LPIPS metric by integrating deep features of classification networks well-trained on ImageNet.

\textbf{Warp Error.} For video colorization, the temporal continuity of generated frames is equally significant with colorization quality. We measure the spatiotemporal consistency by computing the disparity between every warped previous frame and current frame. The warp error of one video is defined as:
\begin{equation}
WE = \sum_{t=2}^T \frac{hw}{hw - sum(M_t)} M_t ||v_t - \mathscr{W} (O_{t-1 \to t}, v_{t-1}) ||_2^2 ,
\end{equation}
where $v_t$ is generated frame at time $t$ and $\mathscr{W} (O_{t-1 \to t}, v_{t-1})$ is the warped frame from previous frame. It is weighted by the number of occlusion pixels. Note that, $M_t$ is a binary mask that considers both occlusion regions and motion boundaries. The $hw$ is the overall number of pixels in a frame. For calculation details, we follow the protocol in \cite{ruder2016artistic}.

%-------------------------------------------------------------------------
\subsection{Video Colorization Comparisons}

We compare VCGAN with existing video colorization algorithms FAVC \cite{lei2019fully}, 3DVC \cite{kouzouglidis2019automatic} and 4 representative image colorization methods CIC \cite{zhang2016colorful}, LTBC \cite{iizuka2016let}, ChromaGAN \cite{vitoria2020chromagan}, and SCGAN \cite{zhao2020scgan}. In addition, we also compare with the task-independent approach BTC \cite{lai2018}, which refines the single image colorization results. Thus, there are 6 video colorization results (\emph{i.e.,} FAVC, 3DVC, CIC + BTC, LTBC + BTC, ChromaGAN + BTC, and SCGAN + BTC) in the experiment. The training sets of all compared methods are the same to VCGAN (\emph{i.e.,} ImageNet \cite{russakovsky2015imagenet}, DAVIS \cite{perazzi2016benchmark}, and Videvo \cite{Videvo}).

\textbf{Qualitative comparison.} Some generated samples from typical methods on two validation sets are shown in Figures \ref{compare1} and \ref{compare2}. On the one hand, the single image colorization methods CIC \cite{zhang2016colorful} and LTBC \cite{iizuka2016let} produce temporally inconsistent results. As shown in the highlighted patches, the colors of objects change extremely, \emph{e.g.,} the sand in Figure \ref{compare1} and the arm in Figure \ref{compare2}. It is because these methods do not consider inter-frame relations. It also demonstrates these image colorization methods are not very robust to shifts or motions. On the other hand, the post-processed results (\emph{i.e.,} CIC + BTC and LTBC + BTC) do not address the above issue obviously. Though BTC \cite{lai2018} induces the temporal relations between every two frames, the colors of image colorization results are ``too different''. BTC cannot handle such cases; therefore, BTC's results are still not continuous enough. In addition, BTC is not jointly trained with single image colorization methods such as CIC and LTBC; therefore, the optimization of image colorization and temporal smoothing are separated. It also causes flickering artifacts in the generated frames. For video colorization algorithms 3DVC \cite{kouzouglidis2019automatic}, FAVC \cite{lei2019fully}, and VCGAN, they do not encounter such issues since temporal correlations are modeled. However, the objects (\emph{e.g.,} the sky, horse, and man) in both Figures \ref{compare1} and \ref{compare2} colorized by FAVC are dusky. The colors of 3DVC results are not natural enough, \emph{e.g.,} the man in Figure \ref{compare2}. Among all the methods, VCGAN produces more colorful and temporally coherent frames. We include more samples and video clips in supplementary materials.

\textbf{Quantitative comparison.} The quantitative comparison on 480p validation sets is concluded in Table \ref{table_sota}. The results of grayscale frames serve as a baseline. Note that we do not include single image colorization methods \cite{zhang2016colorful, iizuka2016let, vitoria2020chromagan, zhao2020scgan} in comparisons since they do not consider the temporal continuity. Firstly, BTC \cite{lai2018} strikes a balance between colorization quality and temporal coherence, the disparity between neighbouring frames is much smaller (\emph{e.g.,} the Warp Error of CIC + BTC is much smaller than CIC). However, the results from CIC + BTC suffer from a decrease of PSNR compared with CIC, since the frame-wise characteristic might be weakened. As discussed, since BTC and CIC are not jointly trained, it also causes decreases in metrics. A similar conclusion also applies to LTBC, ChromaGAN, and SCGAN. Secondly, FAVC \cite{lei2019fully} uses a two-step network, which is jointly optimized by colorization loss function (\emph{e.g.,} L1 loss) and temporal smoothing loss function (\emph{e.g.,} short-term loss). It achieves better PSNR and SSIM results than post-processed results from BTC. Thirdly, 3DVC \cite{kouzouglidis2019automatic} learns both spatial and temporal relations by 3D Conv instead of two-step networks. Therefore, it achieves better PSNR, SSIM, and Warp Error results than FAVC. However, it still adopts a simple 3D U-Net \cite{ronneberger2015u} architecture, which restricts its performances. Finally, VCGAN achieves the best pixel fidelity (PSNR, SSIM) and spatiotemporal consistency (Warp Error) among all the video colorization methods. It demonstrates that the proposed two feature extractors and dense long-term loss $L_{dlt}$ are obviously beneficial to video colorization. The proposed VCGAN uses the semantic model (\emph{i.e.,} two feature extractors), which promotes fast convergence and high colorization quality. In addition, \textbf{the VCGAN is the only hybrid model that unifies both image and video colorization in the same architecture.}

\begin{figure}[t]
\centering
\includegraphics[width=\linewidth]{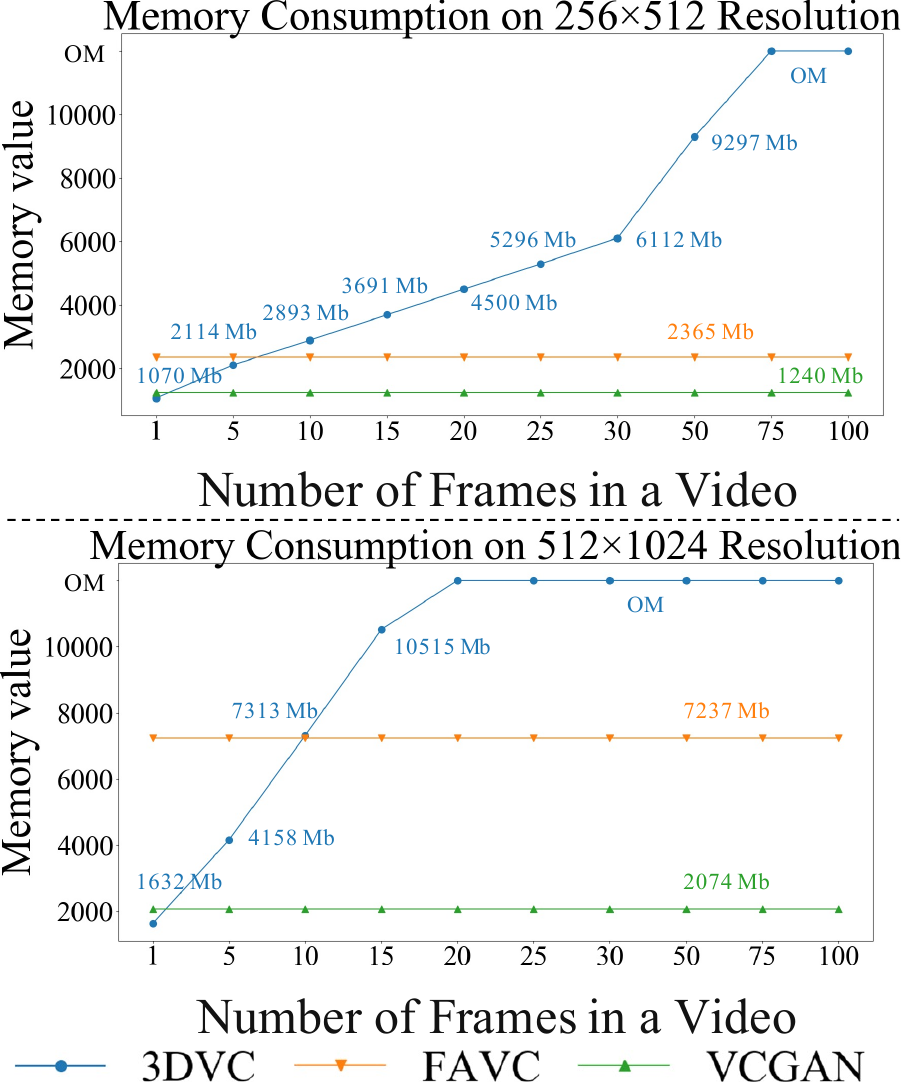}
%, height=10cm

\vspace{-2mm}

\caption{The experiment results on memory consumption. The experiments run on a single NVIDIA Titan Xp GPU with total 12000 Mb memory. ``OM'' denotes ``out of memory'' (\emph{i.e.,} more than 12000 Mb).}
\label{runtime}

\vspace{-0mm}

\end{figure}

\begin{figure*}[t]
\centering
\includegraphics[width=\linewidth, height=12cm]{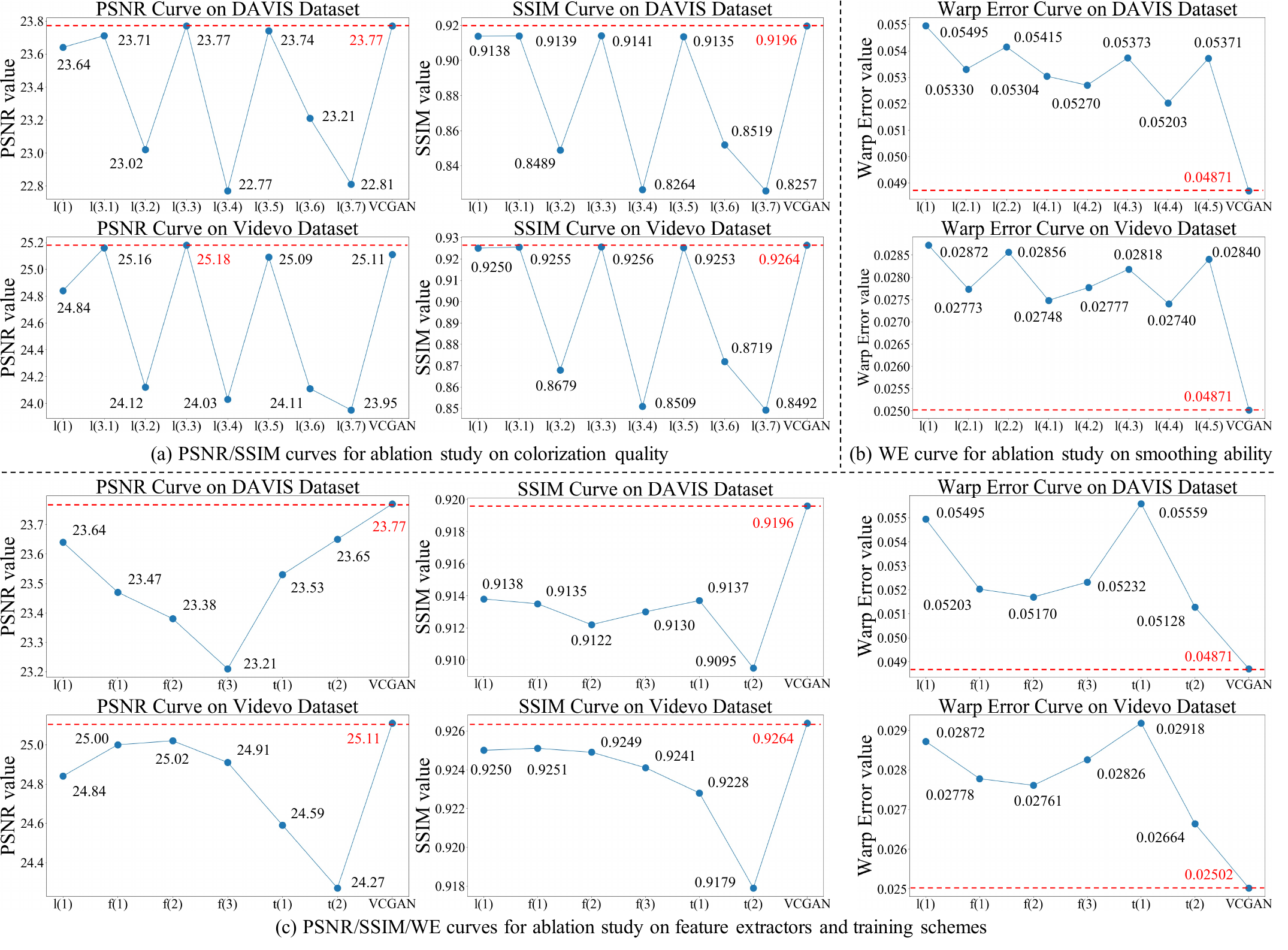}

\vspace{-3mm}

\caption{The quantitative comparisons of ablation study settings. Each sub-figure denotes the results on one metric and one dataset from a group of settings. In each sub-figure, the red value represents the best performances, while the red line contributes to better comparisons.}
\label{abtable}

\vspace{-3mm}

\end{figure*}

%-------------------------------------------------------------------------
\subsection{Memory Analysis}

We conduct a memory analysis for VCGAN and two existing video colorization methods \cite{lei2019fully, kouzouglidis2019automatic} on two image resolutions. The results are concluded in Figure \ref{runtime} and Table \ref{table_macs}. In the experiment, we adjust the number of grayscale frames in a video (to be colorized by the colorization methods) to compute the memory consumption of different methods. The computing platform is one NVIDIA Titan Xp GPU with 12 Gb memory. It is clear that VCGAN has the minimum theoretical MACs and the smallest memory consumption among all the methods. Since 3DVC \cite{kouzouglidis2019automatic} uses 3D Conv, it only ensures the input sequence is temporally related. However, the relations between different sequences are not modeled. Therefore, it easily causes flickering artifacts for long videos due to memory limits. If feeding a longer sequence to 3DVC, it easily encounters ``out of memory'', as shown in Figure \ref{runtime}. FAVC processes two frames simultaneously; however, it has much larger theoretical MACs due to the hyper-column operation. Applying FAVC to color videos still causes much more memory consumption than VCGAN.

\begin{table}[t]
\begin{center}
\caption{Comparison on Multiply–Accumulate Operations (MACs) and GPU Memory per batch (MPB). We select two image resolutions 256$\times$512 (256r) and 512$\times$1024 (512r).}
\label{table_macs}

\vspace{-2mm}

\begin{tabular}{lcccc}
\hline
Method & 256R MACs & 256R MPB & 512R MACs & 512R MPB \\
\hline
\hline
3DVC & 79.72 Mb & OM$>$50 f. & 318.87 Mb & OM$>$15 f. \\
FAVC & 126.69 Mb & 2365 Mb & 506.79 Mb & 7237 Mb \\
VCGAN & \textcolor{red}{60.52 Mb} & \textcolor{red}{1240 Mb} & \textcolor{red}{242.11 Mb} & \textcolor{red}{2074 Mb} \\
\hline
\end{tabular}
\end{center}

\vspace{-8mm}

\end{table}

\begin{table}[t]
\begin{center}
\caption{The conclusion of all ablation study settings, where ``/'' denotes `` no change'' except the last row.}
\label{ab_setting}

\vspace{-2mm}

\begin{tabular}{lccc}
\hline
Setting & Loss terms & FEs & Train scheme  \\
\hline
\hline
l(1) & $L_1$ & / & / \\
l(2.1) & $L_1$, $L_{st}$ & / & / \\
l(2.2) & $L_1$, $L_{dlt}$ & / & / \\
\hline
l(3.1) & $L_p$, $L_G$, $L_{st}$, $L_{dlt}$ & / & / \\
l(3.2) & $L_1$, $L_G$, $L_{st}$, $L_{dlt}$ & / & / \\
l(3.3) & $L_1$, $L_p$, $L_{st}$, $L_{dlt}$ & / & / \\
l(3.4) & $L_1$, $L_{st}$, $L_{dlt}$ & / & / \\
l(3.5) & $L_p$, $L_{st}$, $L_{dlt}$ & / & / \\
l(3.6) & $L_G$, $L_{st}$, $L_{dlt}$ & / & / \\
l(3.7) & $L_{st}$, $L_{dlt}$ & / & / \\
\hline
l(4.1) & $L_1$, $L_p$, $L_G$, $L_{dlt}$ & / & / \\
l(4.2) & $L_1$, $L_p$, $L_G$, $L_{st}$ & / & / \\
l(4.3) & $L_1$, $L_p$, $L_G$ & / & / \\
l(4.4) & $L_1$, $L_p$, $L_G$, $L_{lt}$ & / & / \\
l(4.5) & $L_1$, $L_p$, $L_G$, $L_{st}$, $L_{lt}$ & / & / \\
\hline
f(1) & / & w/o GFE & / \\
f(2) & / & w/o PFE & / \\
f(3) & / & w/o GFE, PFE & / \\
\hline
t(1) & / & / & 1st \\
t(2) & / & / & 2nd, 256p \\
\hline
\scriptsize \textbf{VCGAN} & \textbf{$L_1$, $L_p$, $L_G$, $L_{st}$, $L_{dlt}$} & \textbf{with GFE, PFE} & \textbf{2nd, 480p} \\
\hline
\end{tabular}
\end{center}

\vspace{-5mm}

\end{table}

\begin{figure*}[t]
\centering
\includegraphics[width=\linewidth]{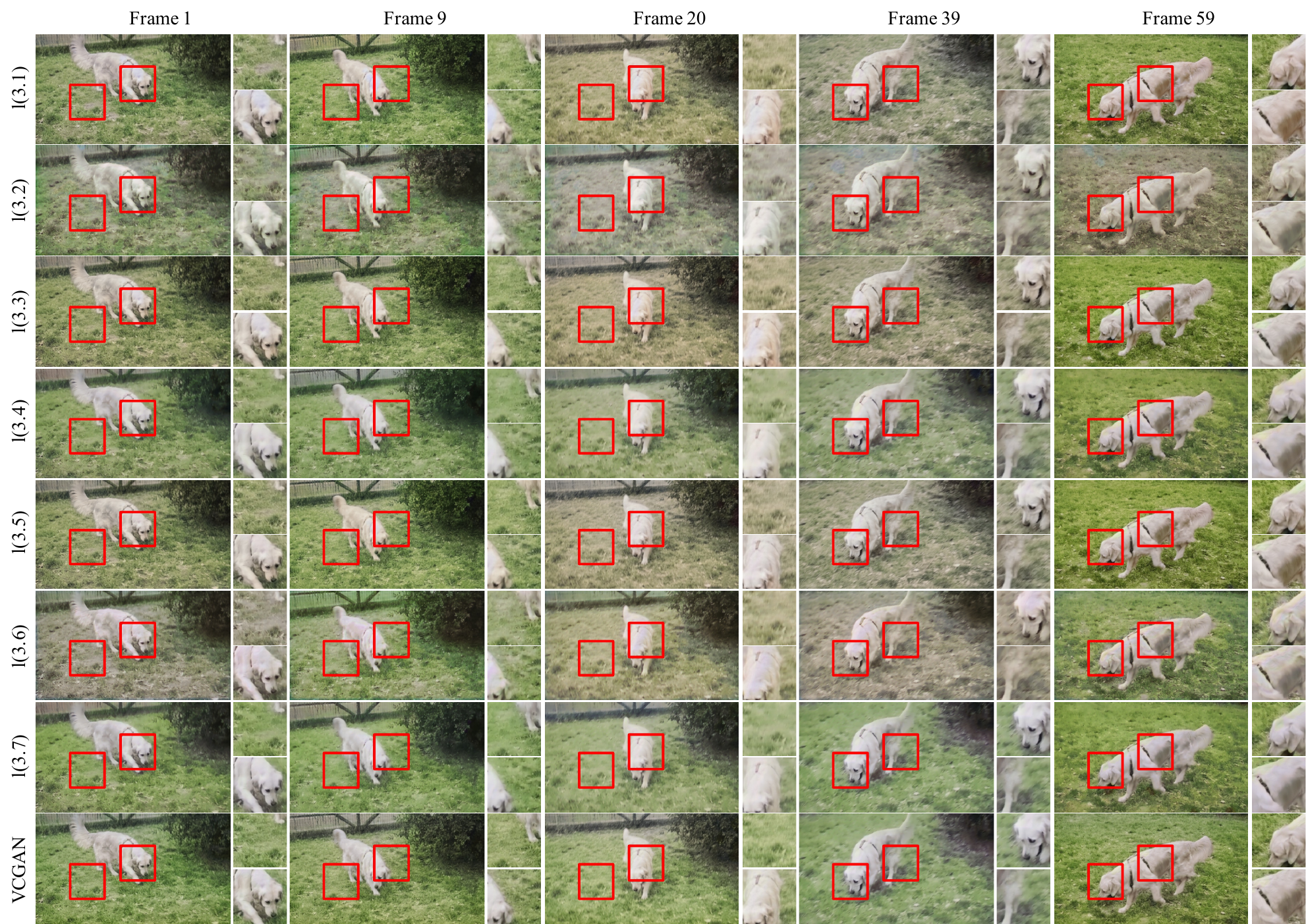}

\vspace{-3mm}

\caption{The comparison of colorization quality on ``dog'' from DAVIS \cite{perazzi2016benchmark} dataset. There are 5 frames shown for 7 ablation study settings. The local patches extracted from the full resolution generated images are placed on the right.}
\label{ablation1}

\vspace{-3mm}

\end{figure*}

\begin{figure*}[t]
\centering
\includegraphics[width=\linewidth, height=10.5cm]{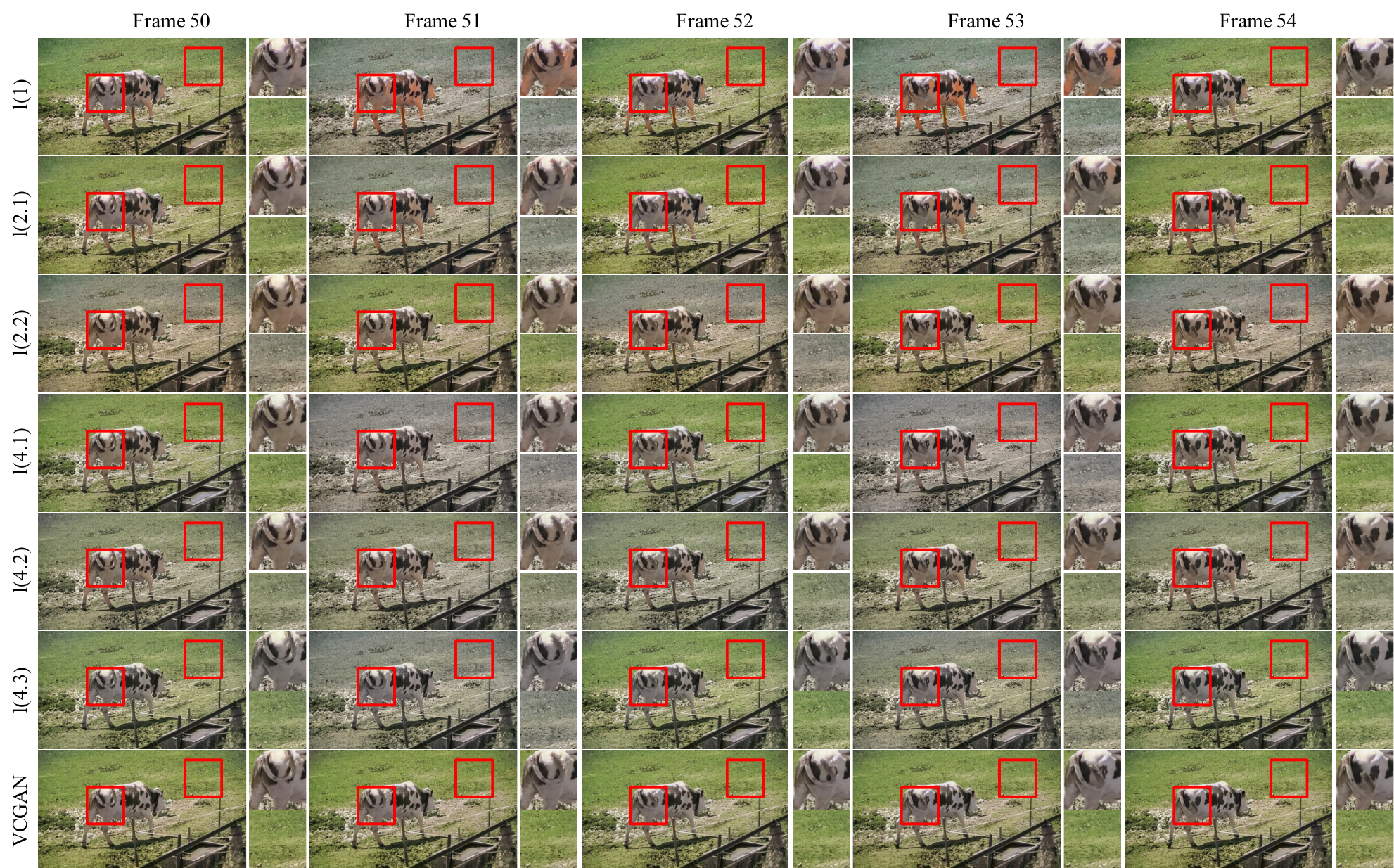}

\vspace{-3mm}

\caption{The comparison of smoothing ability on ``cows'' from DAVIS \cite{perazzi2016benchmark} dataset.}

\vspace{-5mm}

\label{ablation2}
\end{figure*}

\begin{figure*}[t]
\centering
\includegraphics[width=\linewidth]{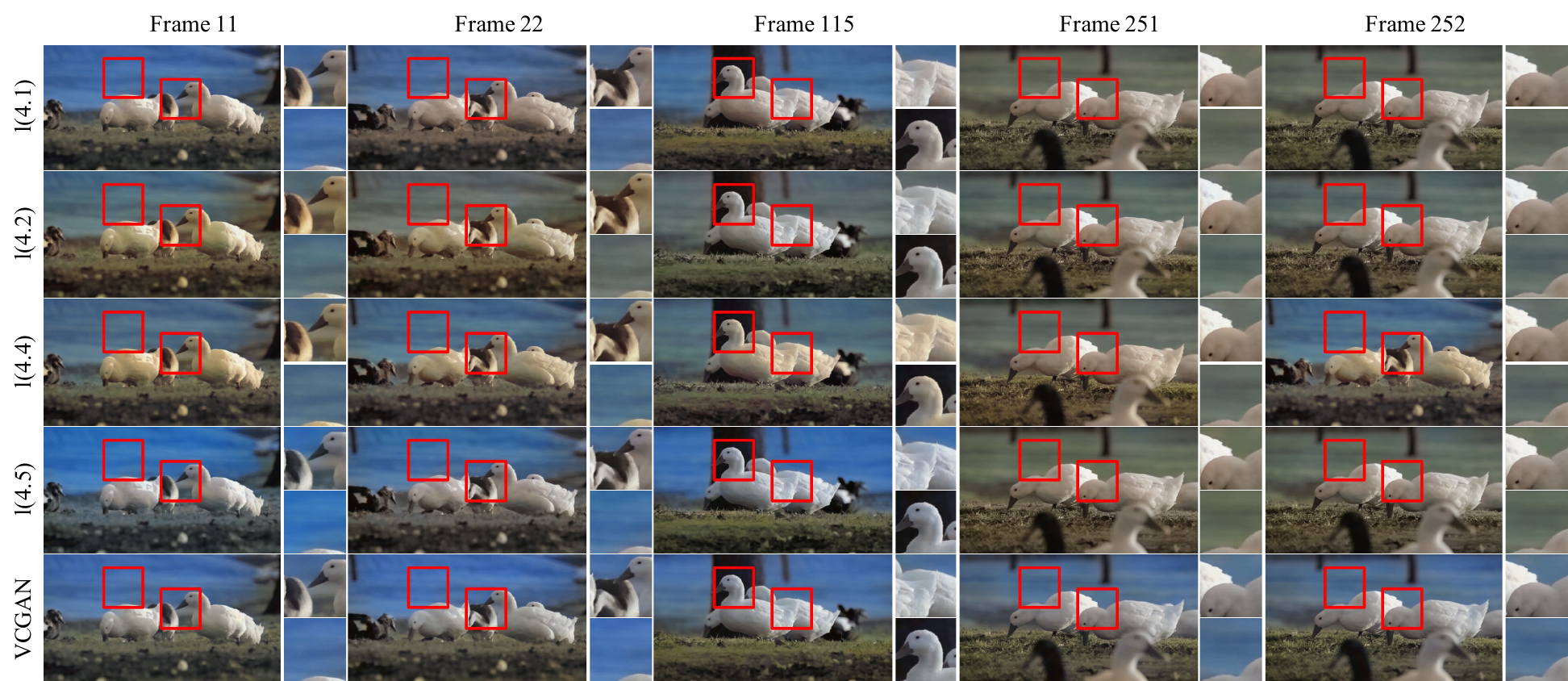}

\vspace{-3mm}

\caption{The comparison of the utility of the proposed dense long-term loss $L_{dlt}$ on ``Ducks'' from Videvo \cite{Videvo} dataset.}

\vspace{-3mm}

\label{ablation3}
\end{figure*}

\begin{figure*}[t]
\centering
\includegraphics[width=\linewidth]{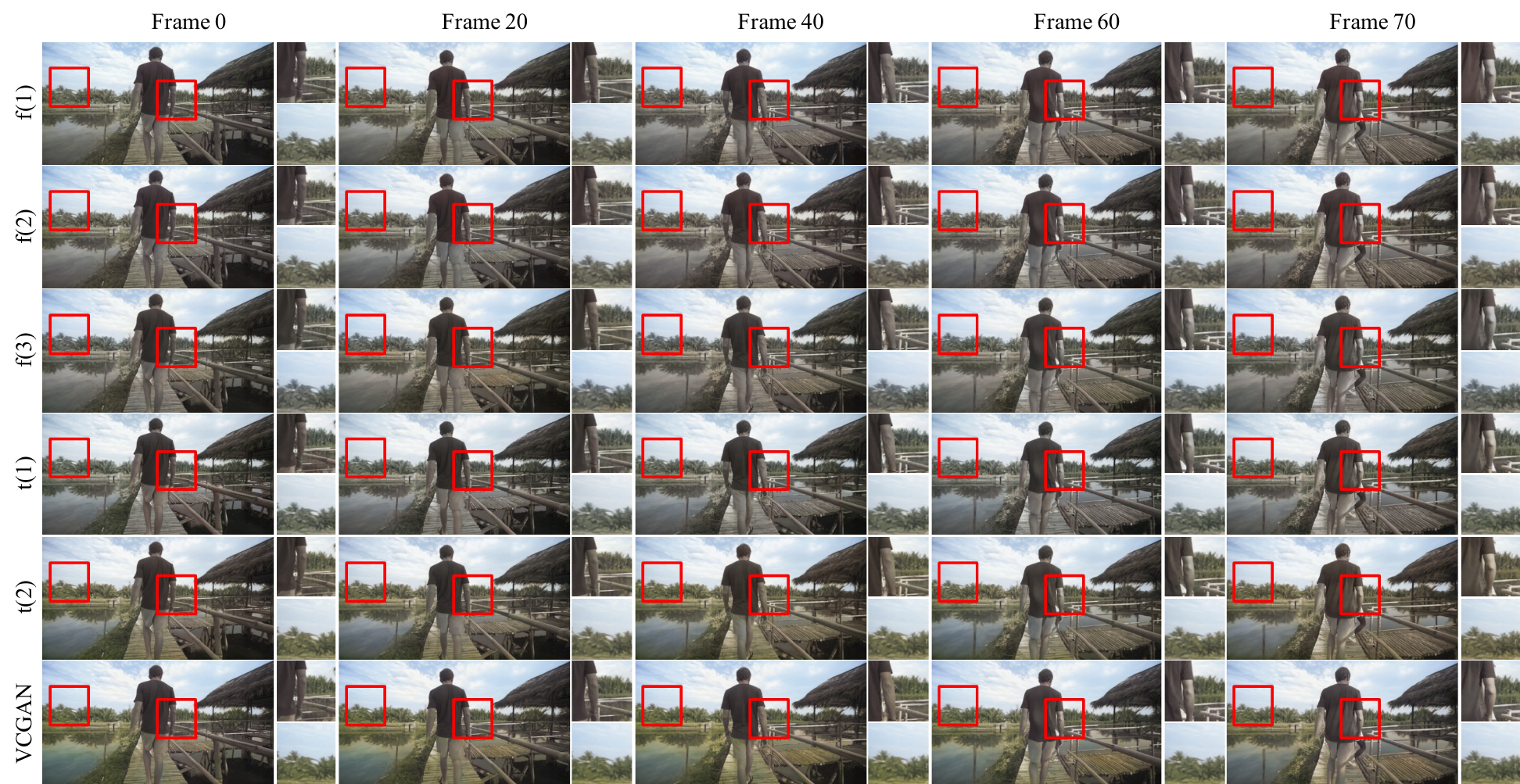}

\vspace{-3mm}

\caption{The comparisons of feature extractors and training schemes on ``YogaHut2'' from Videvo \cite{Videvo} dataset.}

\vspace{-3mm}

\label{ablation4}
\end{figure*}

\begin{figure*}[t]
\centering
\includegraphics[width=\linewidth]{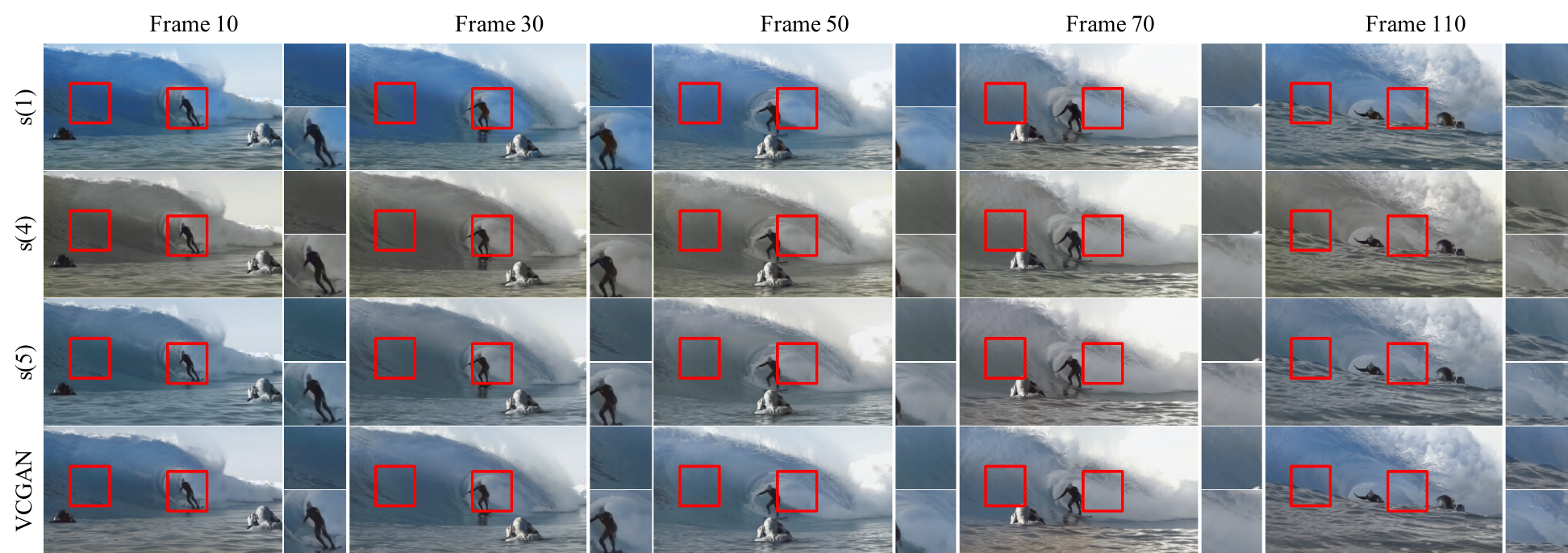}

\vspace{-3mm}

\caption{The comparison of VCGAN trained with different coefficients of loss functions on ``Surfing'' from Videvo \cite{Videvo} dataset.}

\vspace{-5mm}

\label{coeff}
\end{figure*}

%-------------------------------------------------------------------------
\subsection{Ablation Study}

%Since there are many loss terms used in VCGAN, we conduct an ablation study to discover the effectiveness of them on 480p validation data. The training procedure remains unchanged for different settings (\emph{i.e.,} only loss changes). To demonstrate the effectiveness of the proposed training scheme, we include the models of VCGAN first and second training stage (but on 256p training data). The ablation studies are performed on same datasets, \emph{i.e.,} DAVIS \cite{perazzi2016benchmark} and Videvo \cite{Videvo} 480p validation data. There are 16 settings as follows:
To discover the effectiveness of different loss terms, feature extractors, and the proposed training scheme used in VCGAN, We conduct several experiments as an ablation study. The ablation studies are performed on same datasets, \emph{i.e.,} DAVIS \cite{perazzi2016benchmark} and Videvo \cite{Videvo} 480p validation data. There are 20 settings with abbreviations as shown in Table \ref{ab_setting}.

\textbf{Loss terms.} The settings l(1), l(2.1), and l(2.2) serve as baselines. The settings l(3.1)-l(3.7) and l(4.1)-l(4.5) are designed to evaluate ``colorization reality'' (\emph{e.g.,} without $L_1$, $L_p$ and $L_G$) and ``smoothing ability'' (\emph{e.g.,} without $L_{st}$ and $L_{dlt}$), respectively. However, different loss terms have internal relations since the VCGAN is trained by the combinations of the losses with individual coefficients. For instance, if we drop the perceptual loss $L_p$ of VCGAN, the output frames may be smoother than trained with full losses (\emph{i.e.,} Warp Error is smaller). It is because the terms $L_{st}$ and $L_{dlt}$ account relatively more coefficients in this setting than trained with full losses. Thus, we suggest readers \textbf{compare the PSNR and SSIM for ``colorization quality''-related settings (\emph{e.g.,} without $L_p$)} since they may care more about pixel-level accuracy. Similarly, please \textbf{focus on the Warp Error for ``smoothing ability''-related settings (\emph{e.g.,} without $L_{st}$)}. The evaluation results are concluded in Figure \ref{abtable}.

% The setting 1) and 2) (\emph{i.e.,} VCGAN only trained with L1 loss or joint L1 loss and short-term loss) serve as baselines. The other ablation studies can be briefly categorized into two parts in terms of ``colorization reality'' (\emph{e.g.,} without $L_1$, $L_p$ and $L_G$) and ``smoothing ability'' (\emph{e.g.,} without $L_{st}$ and $L_{dlt}$). However, different loss terms have internal relations since the VCGAN is trained by the combination of the losses, using individual coefficients. For instance, if we drop the perceptual loss $L_p$ of VCGAN, the output frames may be smoother than full losses (\emph{i.e.,} Warp Error is smaller). It is because the terms $L_{st}$ and $L_{dlt}$ account more in full loss. Thus, we suggest readers to compare the PSNR and SSIM for ``colorization quality''-related settings (\emph{e.g.,} without $L_p$) since they may care more on pixel-level accuracy. Similarly, please focus on the Warp Error for ``smoothing ability''-related settings (\emph{e.g.,} without $L_{st}$). The setting 8 includes the different training stages of VCGAN so we encourage readers to compare all the metrics. To better represent the point of each ablation setting, we list the evaluation results in Table II.

\textbf{Loss terms related to colorization quality.} As shown in Figure \ref{abtable} (a), the baseline setting l(1) obtains the worst result since it only uses $L_1$ for training. For the settings l(3.1)-l(3.3), if VCGAN trained without $L_1$, $L_p$, or $L_G$, there is a drop in terms of PSNR and SSIM metrics, which demonstrate that all of them are beneficial for colorization quality. As shown in Figure \ref{ablation1}, $L_p$ or $L_G$ promotes VCGAN to generate more realistic and vivid colorizations. Similarly, for the settings l(3.4)-l(3.7), their results are worse than full loss terms, since more than one ``colorization quality''-related loss is removed. Furthermore, the results (\emph{e.g.,} the grass and dog) of l(3.1)-l(3.7) are also of poor color contrast, as shown in Figure \ref{ablation1}, which demonstrates that $L_1$, $L_p$, or $L_G$ are vital for VCGAN to produce high-quality colorizations.
% As shown in Table II, the setting 1) obtains the worst result since it only uses $L_1$ for training. Thus, the colorization is not vivid and continuous frames are not smoothed. By adding the short-term loss $L_{st}$ (setting 2)), the VCGAN has much better results since it meets the Markov Chain assumption. For setting 3), if VCGAN trained without $L_1$, $L_p$ or $L_G$, there is an obvious drop in PSNR and SSIM metrics in both validation set. As an auxiliary loss term, $L_p$ or $L_G$ promotes VCGAN to generate more realistic results. Similarly, for setting 5) and 6), these two metrics are obviously worse than full loss terms. Since more than one ``colorization quality''-related losses are removed, the PSNR and SSIM decrease more than setting 3). However, for setting 3) without $L_{st}$ or $L_{dlt}$ and setting 4) without both, the generated frames are discontinuous, \emph{i.e.,} the Warp Error is bigger than full loss terms. The $L_{st}$ has a bigger effect on minimizing temporal disparity since the neighbouring two frames are closely correlated.

\textbf{Loss terms related to smoothing ability.} As shown in Figure \ref{abtable} (b), setting l(1) obtains the worst Warp Error. By adding the short-term loss $L_{st}$ (setting l(2.1)), VCGAN has better results since it meets the Markov Chain assumption. By adding the dense long-term loss $L_{dlt}$ (setting l(2.2)), VCGAN also performs better due to the consideration of temporal relations. The most significant presumption for video generation is that the produced frames satisfy Markov Chain. Since the setting l(4.1) does not adopt $L_{st}$, it obtains higher Warp Error (\emph{e.g.,} 0.00433 and 0.00246 increases on DAVIS and Videvo, respectively). Similarly, the Warp Error increases if removing $L_{dlt}$ (\emph{i.e.,} l(4.2)) or both $L_{st}$ and $L_{dlt}$ (\emph{i.e.,} l(4.3)). As shown in Figure \ref{ablation2}, the colorized frames from l(4.1)-l(4.3) are not continuous enough, \emph{i.e.,} the colors of continuous frames are not consistent. For the settings l(4.4) and l(4.5), we replace the proposed dense long-term loss $L_{dlt}$ with normal long-term loss $L_{lt}$ \cite{lai2018}, which only panels the differences between current frame and the first frame. The Warp Errors of these settings are still inferior to full VCGAN.

\textbf{Dense long-term loss $L_{dlt}$.} Some previous methods set the time range equals to 2 (previous and current frame) \cite{sheng2013video, xie2018tempogan, lei2019fully, zhang2019deep, kouzouglidis2019automatic} or 3 (previous, current, and leading frames) \cite{ruder2016artistic, kim2019deep}. They did not consider the long-term or remote relations. Lai \emph{et al.} \cite{lai2018} incorporated a long-term loss $L_{lt}$ modeling the connection of current frame and the first frame. However, the proposed dense long-term loss $L_{dlt}$ includes each remote correlation between current frame and all previous generated frames. To demonstrate its effectiveness, we fix the ``colorization quality''-related losses $L_{1}$, $L_{p}$, and $L_{G}$ and use one additional loss from $L_{st}$, $L_{lt}$, and $L_{dlt}$, \emph{i.e.,} l(4.2), l(4.4), and l(4.1). In terms of Warp Error, $L_{st}$ (l(4.1)) is the most significant factor to smooth videos since it panels the neighbouring frames. Though $L_{dlt}$ (l(4.2)) does not contain the consistency of neighbouring frames, it panels each remote frames to minimize the color differences. Compared with $L_{lt}$ (l(4.4)), the $L_{dlt}$ (l(4.2)) achieves lower Warp Error, which demonstrates that modeling all remote relations are beneficial to enhance temporal consistency.

In addition, we add a setting l(4.5) that replaces $L_{dlt}$ with $L_{lt}$. As shown in Figure \ref{abtable} (b) full VCGAN setting, $L_{dlt}$ reduces Warp Errors by approximately 0.00500 and 0.00338 on DAVIS and Videvo datasets, respectively, than $L_{lt}$. In addition, since the continuous frames may not represent long-term consistency, we illustrate remote frames in Figure \ref{ablation3} to show the effect of the proposed dense long-term loss $L_{dlt}$. Only the VCGAN trained with $L_{dlt}$ produces consistent background color (\emph{i.e.,} blue sky); whereas the normal long-term loss $L_{lt}$ fails to maintain the consistency for remote frames. In all settings, VCGAN with full losses better balances colorization fidelity and spatiotemporal constancy. However, other settings will one-sidedly emphasize PSNR or warp error, which demonstrates each loss term is significant for VCGAN.

% The most significant presumption for video generation is that the produced frames satisfy Markov Chain. Some previous methods set the time range equals to 2 (previous and current frame) \cite{sheng2013video, xie2018tempogan, lei2019fully, zhang2019deep, kouzouglidis2019automatic} or 3 (previous, current, and leading frames) \cite{ruder2016artistic, kim2019deep}. They do not consider the long-term relationship. Lai \emph{et al.} \cite{lai2018} incorporate a long-term loss function modelling connection of current frame and first frame. However, the proposed dense long-term consistency includes each remote correlation and requires ignorable additional memory. We compare the proposed dense long-term loss and traditional long-term loss \cite{lai2018} (setting 7)). The dense long-term loss explicitly enhances warp error by approximately 0.4 than normal long-term loss, which is essential for maintaining spatiotemporal consistency in video generation tasks. Also, it promotes near frames more similar while maintains the colorization quality. In all settings, VCGAN with full losses better balance colorization fidelity and spatiotemporal constancy. However, other settings will one-sidedly emphasize PSNR or warp error, which demonstrate each loss term is significant for VCGAN.

\begin{table*}[t]
\begin{center}
\caption{The experiment conclusion of the sensitiveness of loss coefficients. The \textcolor{red}{\textbf{red}}, \textcolor{blue}{\textbf{blue}}, and \textcolor{green}{green} colors represent the best, the second-best, and the third-best performances, respectively.}
\label{loss_setting}

\begin{tabular}{lccccclcccccc}
\hline
\multirow{2}{*}{Setting} & \multirow{2}{*}{$\lambda_1$} & \multirow{2}{*}{$\lambda_p$} & \multirow{2}{*}{$\lambda_G$} & \multirow{2}{*}{$\lambda_{st}$} & \multirow{2}{*}{$\lambda_{dlt}$} & \multirow{2}{*}{Target} & \multicolumn{3}{c}{DAVIS} & \multicolumn{3}{c}{Videvo} \cr & & & & & & & PSNR & SSIM & Warp Error & PSNR & SSIM & Warp Error \\
\hline
\hline
s(1) & 1 & 1 & 1 & 1 & 1 & all ``1'' coefficients & \textcolor{green}{\textbf{23.83}} & \textcolor{green}{\textbf{0.9193}} & 0.05101 & 24.68 & 0.9224 & 0.02659 \\
s(2) & 20 & 5 & 1 & 3 & 5 & double $\lambda_1$ & \textcolor{red}{\textbf{23.90}} & 0.9192 & 0.05042 & \textcolor{blue}{\textbf{25.11}} & \textcolor{green}{\textbf{0.9244}} & 0.02746 \\
s(3) & 10 & 10 & 1 & 3 & 5 & double $\lambda_p$ & \textcolor{blue}{\textbf{23.85}} & \textcolor{red}{\textbf{0.9202}} & 0.04971 & \textcolor{red}{\textbf{25.20}} & 0.9232 & 0.02602 \\
s(4) & 10 & 5 & 2 & 3 & 5 & double $\lambda_G$ & 23.32 & 0.9113 & 0.04957 & 24.66 & 0.9211 & 0.02644 \\
s(5) & 10 & 5 & 1 & 6 & 5 & double $\lambda_{st}$ & 23.75 & 0.9133 & \textcolor{green}{\textbf{0.04915}} & 24.55 & 0.9197 & 0.02565 \\
s(6) & 10 & 5 & 1 & 3 & 10 & double $\lambda_{dlt}$ & 23.37 & 0.9096 & \textcolor{blue}{\textbf{0.04909}} & 24.67 & 0.9194 & \textcolor{green}{\textbf{0.02536}} \\
s(7) & 20 & 10 & 2 & 3 & 5 & double $\lambda_1$, $\lambda_p$, and $\lambda_G$ & 23.63 & 0.9118 & 0.04933 & \textcolor{green}{\textbf{25.07}} & \textcolor{blue}{\textbf{0.9245}} & 0.02650 \\
s(8) & 10 & 5 & 1 & 6 & 10 & double $\lambda_{st}$ and $\lambda_{dlt}$ & 23.76 & 0.9127 & \textcolor{red}{\textbf{0.04871}} & 24.56 & 0.9199 & \textcolor{red}{\textbf{0.02501}} \\
\hline
\textbf{VCGAN} & \textbf{10} & \textbf{5} & \textbf{1} & \textbf{3} & \textbf{5} & \textbf{full VCGAN} & 23.77 & \textcolor{blue}{\textbf{0.9196}} & \textcolor{red}{\textbf{0.04871}} & \textcolor{blue}{\textbf{25.11}} & \textcolor{red}{\textbf{0.9264}} & \textcolor{blue}{\textbf{0.02502}} \\
\hline
\end{tabular}
\end{center}

\vspace{-5mm}

\end{table*}

\textbf{Feature extractors.} To demonstrate the advance of the proposed two feature extractors, we remove the global feature extractor (GFE) or placeholder feature extractor (PFE) or both for comparisons (\emph{i.e.,} f(1), f(2), and f(3)). The GFE is a pre-trained ResNet-50-IN, which provides semantics for the VCGAN to identify colors for objects with similar edges \cite{zhao2020scgan}. Therefore, f(1) obtains worse PSNR and SSIM values. Also, we found the Warp Errors of f(1) are higher than full VCGAN, which proves that the semantics provided by the pre-trained GFE are also beneficial to minimize inter-frame disparity. For f(2), it proves that the PFE can provide the information from last colorized frame. Otherwise, the Warp Error increases due to no use of the PFE with recurrent connection. For f(3), it obtains worse results since only the mainstream of VCGAN is used. As shown in Figure \ref{ablation4}, the patches are less colorful than full VCGAN.

\textbf{Training scheme.} For the proposed training scheme, we include the VCGAN first and second training stage (on 256p resolution) models for comparisons (\emph{i.e.,} t(1) and t(2)). Since the image resolution and loss terms (\emph{e.g.,} temporal losses) are both different from the full VCGAN, directly applying first stage model leads to extremely inconsistent videos. Similarly, if the training resolution and testing resolution are unequal, the result is not plausible. Some results are shown in Figure \ref{ablation4}, where the colors are not vivid enough and the frames are not continuous enough compared with the full VCGAN.

In conclusion, each component is vital for the proposed VCGAN to obtain high-quality and temporally smooth video colorizations. Also, the proposed dense long-term loss further ensures the consistency of far frames.

%-------------------------------------------------------------------------

\subsection{Investigation of the Sensitiveness of Loss Coefficients}

The coefficients of the objectives used for VCGAN optimization are empirically selected. To demonstrate that the proposed coefficient combination is relatively better than other combinations, we conduct several experiments by adjusting some of the coefficients. The results on DAVIS and Videvo datasets are in Table \ref{loss_setting}. The proposed coefficients achieve relatively better values in terms of PSNR, SSIM, and Warp Error metrics. Also as shown in Figure \ref{coeff}, if VCGAN trained with all ``1'' coefficients (\emph{i.e.,} s(1)), the colors are very consistent for far frames, since it may not balance high-quality colorization and temporal consistency well. If doubling $\lambda_G$ (\emph{i.e.,} s(4)), the results are almost monochrome. If doubling $\lambda_{sl}$ (\emph{i.e.,} s(5)), the colors are also less vivid. In conclusion, the proposed coefficient combination is relatively better.

\subsection{Image Colorization Results}

If the input for the placeholder feature extractor is replaced with a grayscale image, the proposed VCGAN turns into an image colorization model (\emph{i.e.,} a video only contains one frame). To demonstrate the image colorization ability of VCGAN, we compare the VCGAN first training stage model with 7 state-of-the-art image colorization algorithms \cite{zhang2016colorful, iizuka2016let, isola2017image, DeOldify, lei2019fully, vitoria2020chromagan}, where the colorization part of \cite{lei2019fully} is adopted. Note that, the training sets of the methods are the same (\emph{i.e.,} ImageNet \cite{russakovsky2015imagenet}). Following the settings in \cite{zhang2016colorful}, we choose the 10000 images from the ImageNet validation set for evaluation.

\begin{figure*}[t]
\centering
\includegraphics[width=\linewidth]{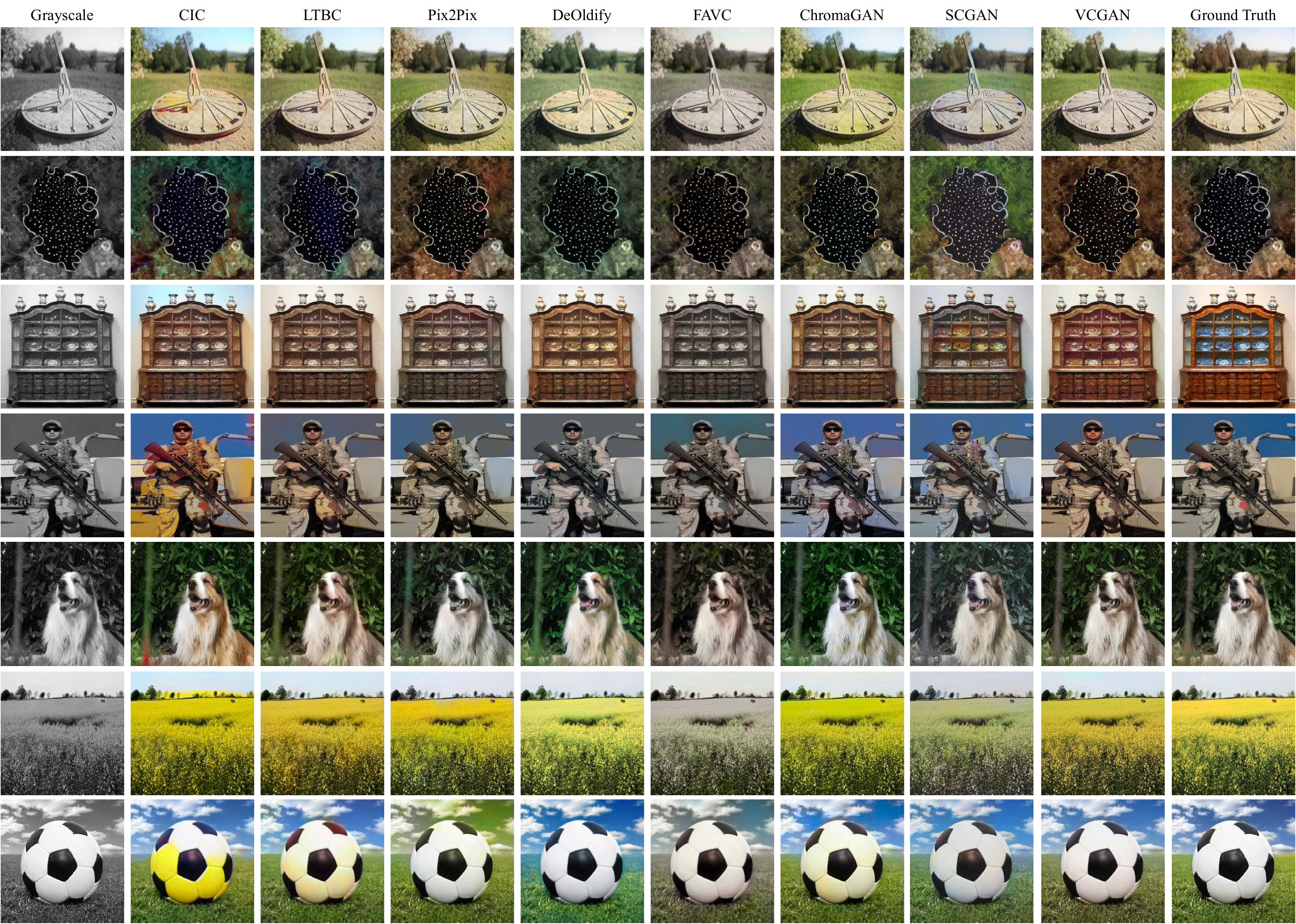}

\vspace{-2mm}

\caption{Illustration of image colorization results of VCGAN (first stage) and state-of-the-art methods \cite{zhang2016colorful, iizuka2016let, isola2017image, DeOldify, lei2019fully, vitoria2020chromagan, zhao2020scgan} on ImageNet validation set. The first column and last column denote the grayscale and colorful ground truth. The other columns include the colorizations of the methods in the experiment. The red rectangles in the figures represent inconsistent regions or strange colors.}

\vspace{-0mm}

\label{image}
\end{figure*}

We illustrate some colorized results in Figure \ref{image}. There are obvious visual artifacts in the generated results of other methods. For instance, there are color bleeding artifacts (\emph{i.e.,} the color of one object permeates to other objects) in rows 1-4 of CIC \cite{zhang2016colorful} and rows 3 and 4 of DeOldify \cite{DeOldify}. Even though there are no color bleeding artifacts of FAVC \cite{lei2019fully}, their results are not colorful enough compared with other methods. However, the results generated by VCGAN are more colorful and reasonable than other methods. Also, there are almost no artifacts in the results of VCGAN. In conclusion, the hybrid VCGAN architecture is appropriate for both image and video colorization tasks.

The quantitative analysis is summarized in Table \ref{table_image}. The proposed VCGAN achieves the best PSNR. It demonstrates that the VCGAN architecture produces the colorizations with the highest pixel fidelity. Also, it obtains the second-best SSIM and Top-5 Accuracy metrics, which evaluate the semantic representation ability of colorization systems. It demonstrates that the VCGAN architecture can generate relatively more plausible colorizations than other methods. The GAN-based methods (Pix2Pix \cite{isola2017image}, DeOldify \cite{DeOldify}, ChromaGAN \cite{vitoria2020chromagan} and the proposed VCGAN) obtain better performance, since the GAN facilitates sharper results, which are difficult to accomplish by only adopting L1 loss.

\begin{table}[t]
\begin{center}
\caption{Comparison of state-of-the-art image colorization methods \cite{zhang2016colorful, iizuka2016let, isola2017image, DeOldify, lei2019fully, vitoria2020chromagan, zhao2020scgan} and proposed VCGAN (first stage). The \textcolor{red}{\textbf{red}}, \textcolor{blue}{\textbf{blue}}, and \textcolor{green}{green} colors represent the best, the second-best, and the third-best performances, respectively.}
\label{table_image}

\vspace{-2mm}

\begin{tabular}{lcccc}
\hline
Method & PSNR & SSIM & Top-5 Acc & GAN Training \\
\hline
\hline
Ground Truth & / & 1 & 84.91\% & / \\
Grayscale & 23.23 & 0.9394 & 73.81\% & / \\
\hline
CIC \cite{zhang2016colorful} & 22.49 & 0.9153 & \textcolor{green}{\textbf{78.11\%}} & / \\
LTBC \cite{iizuka2016let} & \textcolor{blue}{\textbf{24.32}} & \textcolor{blue}{\textbf{0.9464}} & 77.13\% & / \\
Pix2Pix \cite{isola2017image} & 23.25 & 0.9386 & 76.57\% & \checkmark \\
DeOldify \cite{DeOldify} & 23.14 & 0.9194 & 78.01\% & \checkmark \\
FAVC \cite{lei2019fully} & 22.96 & 0.9146 & 76.76\% & / \\
ChromaGAN \cite{vitoria2020chromagan} & \textcolor{blue}{\textbf{24.32}} & 0.9273 & \textcolor{red}{\textbf{78.51\%}} & \checkmark \\
SCGAN \cite{zhao2020scgan} & \textcolor{green}{\textbf{23.80}} & \textcolor{red}{\textbf{0.9470}} & 76.70\% & \checkmark \\
VCGAN & \textcolor{red}{\textbf{24.48}} & \textcolor{green}{\textbf{0.9427}} & \textcolor{blue}{\textbf{78.19\%}} & \checkmark \\
\hline
\end{tabular}
\end{center}

\vspace{-0mm}

\end{table}

\begin{figure*}[t]
\centering
\includegraphics[width=\linewidth]{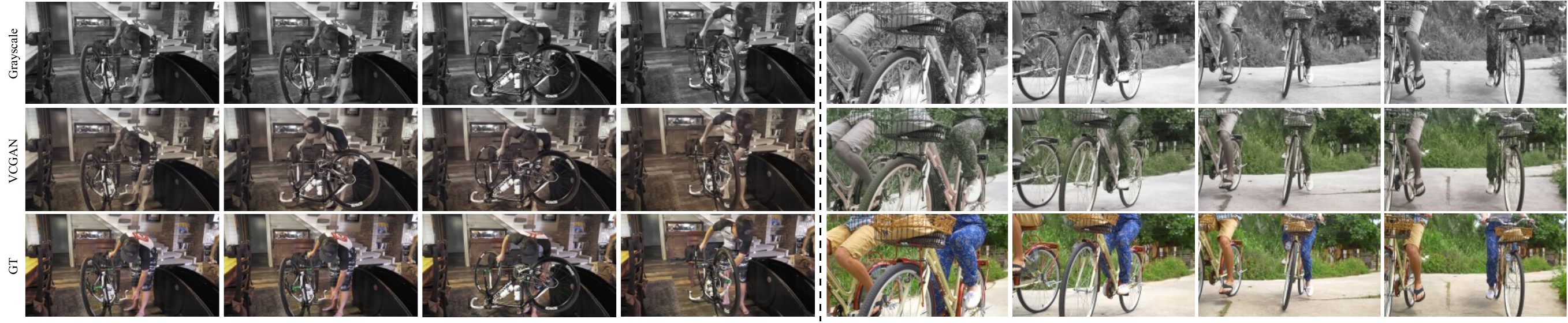}

\vspace{-2mm}

\caption{Failure cases of VCGAN. The rows from top to bottom denote the grayscale input, colorized frames by VCGAN, and ground truth, respectively.}

\vspace{-0mm}

\label{failure}
\end{figure*}

%-------------------------------------------------------------------------
\subsection{Failure Cases}

The proposed VCGAN produces relatively plausible colorful videos in many cases. However, there still exists a common issue when there are a lot of details in each frame (please refer to the left part of Figure \ref{failure}). Also, since the video colorization is an ill-posed problem, the produced frames might be not colorful enough (please refer to the right part of Figure \ref{failure}). The more complicated video training datasets may enhance the performance of VCGAN. In the future, we will further improve VCGAN architecture to make it faster and produce more plausible and colorful results.

%------------------------------------------------------------------------
\section{Conclusion}

In this paper, we presented a recurrent VCGAN framework to automatically generate photorealistic and temporally coherent video colorization. Utilizing two pre-trained ResNet-50-IN networks as the global feature extractor and placeholder feature extractor along with the U-Net-based mainstream, the VCGAN generator extracts semantics efficiently while maintains the spatiotemporal consistency among consecutive frames recurrently. By changing the input for placeholder feature extractor, VCGAN architecture unifies both image and video colorization applications. Furthermore, the proposed dense long-term loss models every remote relations for far frames. It enhances the smoothness of generated videos while requires ignorable additional memory. The adversarial loss is also adopted in the video colorization domain to improve the color vividness. Finally, we validated VCGAN with several state-of-the-art image and video colorization methods. The experiment shows that VCGAN has the minimum theoretical MACs and the smallest memory consumption among current video colorization methods. The experiment also demonstrates that the proposed VCGAN obtains better performances in both image and video colorization applications than the other well-known methods.

% if have a single appendix:
%\appendix[Proof of the Zonklar Equations]
% or
%\appendix  % for no appendix heading
% do not use \section anymore after \appendix, only \section*
% is possibly needed

% use appendices with more than one appendix
% then use \section to start each appendix
% you must declare a \section before using any
% \subsection or using \label (\appendices by itself
% starts a section numbered zero.)
%

%\appendices
%\section{Proof of the First Zonklar Equation}
%Appendix one text goes here.
%
%% you can choose not to have a title for an appendix
%% if you want by leaving the argument blank
%\section{}
%Appendix two text goes here.

% use section* for acknowledgment
\section*{Acknowledgment}

The authors would like to thank Bei Li, Pengfei Xian, Xuihui Wang and Wei Liu for many helpful comments. The authors would also like to thank the anonymous reviewers and the editors for their kind suggestions.

%This work was supported by an Internal Funds Scheme from City University of Hong Kong under Project 9678141.

% Can use something like this to put references on a page
% by themselves when using endfloat and the captionsoff option.
\ifCLASSOPTIONcaptionsoff
  \newpage
\fi

% trigger a \newpage just before the given reference
% number - used to balance the columns on the last page
% adjust value as needed - may need to be readjusted if
% the document is modified later
%\IEEEtriggeratref{8}
% The "triggered" command can be changed if desired:
%\IEEEtriggercmd{\enlargethispage{-5in}}

% references section

% can use a bibliography generated by BibTeX as a .bbl file
% BibTeX documentation can be easily obtained at:
% http://mirror.ctan.org/biblio/bibtex/contrib/doc/
% The IEEEtran BibTeX style support page is at:
% http://www.michaelshell.org/tex/ieeetran/bibtex/
%\bibliographystyle{IEEEtran}
% argument is your BibTeX string definitions and bibliography database(s)
%\bibliography{IEEEabrv,../bib/paper}
%
% <OR> manually copy in the resultant .bbl file
% set second argument of \begin to the number of references
% (used to reserve space for the reference number labels box)

{
\bibliographystyle{IEEEtran}
\bibliography{IEEEabrv,mybibfile}
}

%\begin{thebibliography}{1}
%
%\bibitem{IEEEhowto:kopka}
%H.~Kopka and P.~W. Daly, \emph{A Guide to \LaTeX}, 3rd~ed.\hskip 1em plus
%  0.5em minus 0.4em\relax Harlow, England: Addison-Wesley, 1999.
%
%\end{thebibliography}

% biography section
% 
% If you have an EPS/PDF photo (graphicx package needed) extra braces are
% needed around the contents of the optional argument to biography to prevent
% the LaTeX parser from getting confused when it sees the complicated
% \includegraphics command within an optional argument. (You could create
% your own custom macro containing the \includegraphics command to make things
% simpler here.)
%\begin{IEEEbiography}[{\includegraphics[width=1in,height=1.25in,clip,keepaspectratio]{mshell}}]{Michael Shell}
% or if you just want to reserve a space for a photo:

\vspace{0cm}

\begin{IEEEbiography}[{\includegraphics[width=1in,height=1.25in,clip,keepaspectratio]{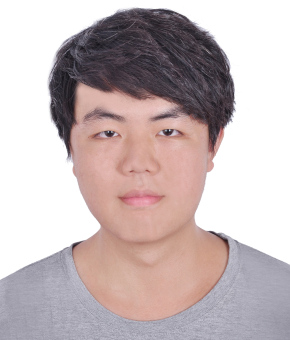}}]{Yuzhi Zhao}

(S’19) received the B.Eng. Degree in electronic information from Huazhong University of Science and Technology, Wuhan, China, in 2018. He is currently pursuing the Ph.D. degree with the Department of Electronic Engineering, City University of Hong Kong. His research interests include image processing, deep learning, and machine learning.

\end{IEEEbiography}

% if you will not have a photo at all:
\begin{IEEEbiography}[{\includegraphics[width=1in,height=1.25in,clip,keepaspectratio]{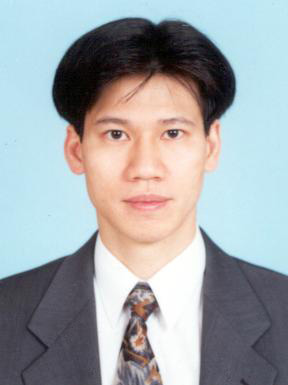}}]{Lai-Man Po}

(M’92–SM’09) received the B.S. and Ph.D. degrees in electronic engineering from the City University of Hong Kong, Hong Kong, in 1988 and 1991, respectively. He has been with the Department of Electronic Engineering, City University of Hong Kong, since 1991, where he is currently an Associate Professor of Department of Electronic Engineering. He has authored over 150 technical journal and conference papers. His research interests include image and video coding with an emphasis deep learning based computer vision algorithms.

Dr. Po is a member of the Technical Committee on Multimedia Systems and Applications and the IEEE Circuits and Systems Society. He was the Chairman of the IEEE Signal Processing Hong Kong Chapter in 2012 and 2013. He was an Associate Editor of HKIE Transactions in 2011 to 2013. He also served on the Organizing Committee, of the IEEE International Conference on Acoustics, Speech and Signal Processing in 2003, and the IEEE International Conference on Image Processing in 2010.

\end{IEEEbiography}

\begin{IEEEbiography}[{\includegraphics[width=1in,height=1.25in,clip,keepaspectratio]{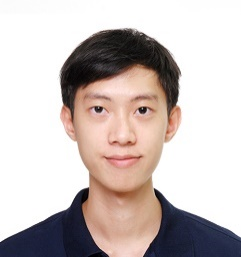}}]{Wing-Yin Yu}

(S’21) received the B.Eng. degree in Information Engineering from City University of Hong Kong, in 2019. He is currently pursuing the Ph.D. degree at Department of Electronic Engineering at City University of Hong Kong. His research interests are deep learning and computer vision.

\end{IEEEbiography}

\begin{IEEEbiography}[{\includegraphics[width=1in,height=1.7in,clip,keepaspectratio]{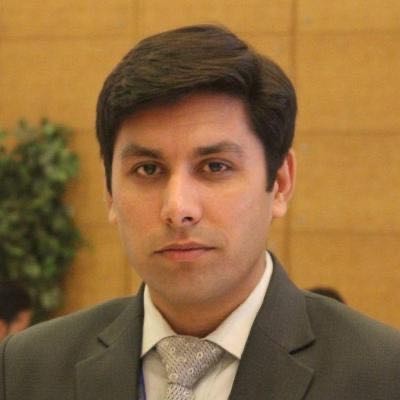}}]{Yasar Abbas Ur Rehman}

(S’19–M’20) received the B.Sc. degree in electrical engineering (telecommunication) from the City University of Science and Information Technology, Peshawar, Pakistan, in 2012, the M.Sc. degree in electrical engineering from the National University of Computer and Emerging Sciences, Pakistan, in 2015, and Ph.D. degree in Electronic Engineering from City University of Hong Kong, Hong Kong, in 2019. He is currently working with TCL corporate research (HK) Co., Ltd as a postdoctoral researcher. His research interests include computer vision, machine learning, deep learning and its applications in facial recognition, biometric anti-spoofing, and video understanding.

\end{IEEEbiography}

\begin{IEEEbiography}[{\includegraphics[width=1in,height=1.7in,clip,keepaspectratio]{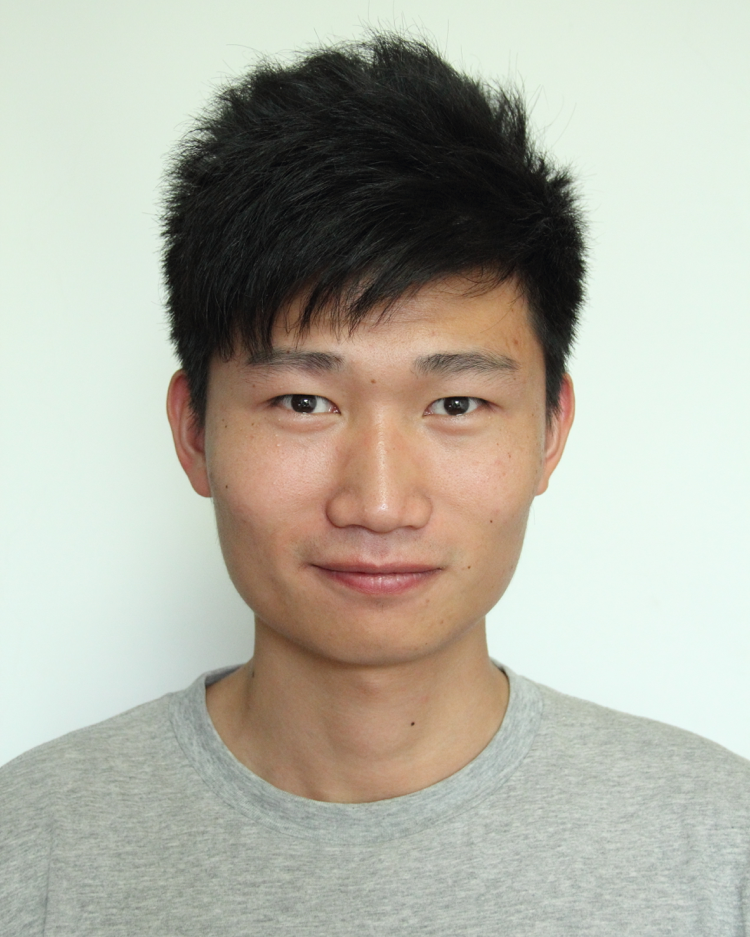}}]{Mengyang Liu}

received the B.Eng. degree in optoelectronic engineering from the Shanghai University of Electric Power, Shanghai, China, in 2014, and the M.Sc. degree in electronic and information engineering and the Ph.D. degree from the City University of Hong Kong, in 2015 and 2019, respectively. He is currently an Engineer with the Tencent Video, Tencent Holdings Ltd. His research interests include image and video processing, video embedding and retrieval, computer vision, and machine learning.

\end{IEEEbiography}

\begin{IEEEbiography}[{\includegraphics[width=1in,height=1.7in,clip,keepaspectratio]{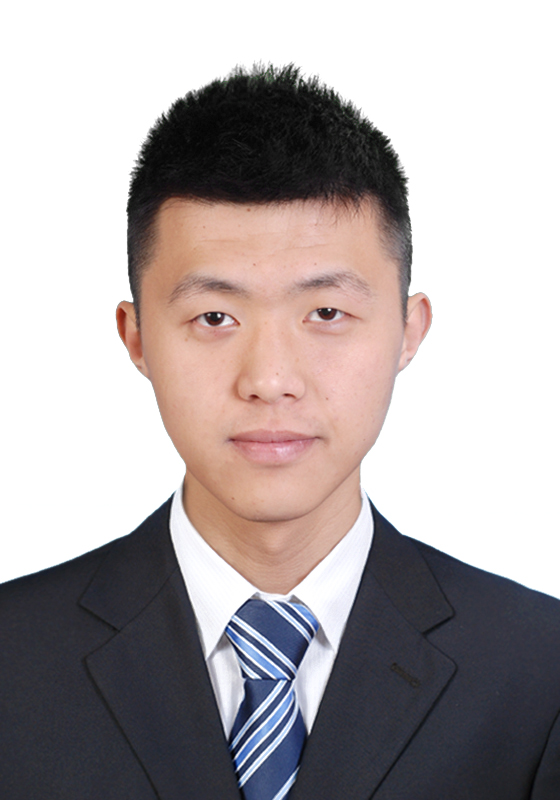}}]{Yujia Zhang}

received the B.Eng. degree in electrical engineering and automation in Huazhong University of Science and Technology in 2015, and the M.Sc. degree in electrical engineering in South China University of Technology, China, in 2018. He is currently pursuing the Ph.D. degree in City University of Hong Kong. His current research interests include computer vision and video understanding.

\end{IEEEbiography}

\begin{IEEEbiography}[{\includegraphics[width=1in,height=1.7in,clip,keepaspectratio]{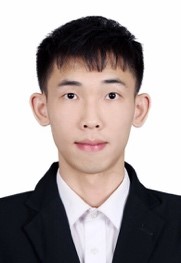}}]{Weifeng Ou}

received his B.Eng. degree in Telecommunication Engineering from Guangdong University of Technology in 2013, his M.Eng. degree in Signal \& Information Processing from South China University of Technology in 2016, and his Ph.D. degree in the Department of Electrical Engineering from City University of Hong Kong in 2021. He was with Huawei as an R \& D engineer from 2016 to 2018. He is currently working in SenseTime Group Limited. His research interests include biometrics and deep learning. 

\end{IEEEbiography}

% insert where needed to balance the two columns on the last page with
% biographies
%\newpage
%
%\begin{IEEEbiographynophoto}{Jane Doe}
%Biography text here.
%\end{IEEEbiographynophoto}

% You can push biographies down or up by placing
% a \vfill before or after them. The appropriate
% use of \vfill depends on what kind of text is
% on the last page and whether or not the columns
% are being equalized.

%\vfill

% Can be used to pull up biographies so that the bottom of the last one
% is flush with the other column.
%\enlargethispage{-5in}

% that's all folks
\end{document}